%% file: arXiv.tex
\documentclass{article}

\usepackage[preprint]{neurips_2022}

\usepackage[utf8]{inputenc} 
\usepackage[T1]{fontenc}    
\usepackage{hyperref}       
\usepackage{url}            
\usepackage{booktabs}       
\usepackage{amsfonts}       
\usepackage{nicefrac}       
\usepackage{microtype}      
\usepackage{xcolor}         

\usepackage{tikz}
\usepackage{comment}
\usepackage{amsmath,amssymb} 
\usepackage{color}
\usepackage{graphicx}
\usepackage{amsmath}
\usepackage{amssymb}
\usepackage{mathrsfs}
\usepackage{algorithm}
\usepackage{algpseudocode}
\usepackage{xspace}
\usepackage{wrapfig}
\usepackage{enumitem}

\input{math_commands.tex}

\usepackage{xcolor}

\newcommand{\etal}{\textit{et al}.}

\usepackage[accsupp]{axessibility}  

\usepackage{enumitem}

\title{Structured Graph Variational Autoencoders for Indoor Furniture layout Generation}

\author{%
  Aditya Chattopadhyay \thanks{This work was done during Aditya's internship with Amazon}\\
  Department of Computer Science\\
  Johns Hopkins University\\
  \texttt{achatto1@jhu.edu} \\
  \And
  Xi Zhang \\
  Amazon.com, Inc. \\
  \texttt{xizhn@amazon.com} \\
  \AND
  David Paul Wipf \\
  Amazon.com, Inc. \\
  \texttt{daviwipf@amazon.com} \\
  \And
  Himanshu Arora \\
  Amazon.com, Inc. \\
  \texttt{arorah@amazon.com} \\
  \And
  Ren\'e Vidal \\
  Amazon.com, Inc. \\
  \texttt{rvidal@amazon.com} \\
}

\begin{document}

\maketitle

\begin{abstract}
We present a graph variational autoencoder with a structured prior for generating the layout of indoor 3D scenes. Given the room type (e.g., living room or library) and the room layout (e.g., room elements such as floor and walls), our architecture generates a collection of objects (e.g., furniture items such as sofa, table and chairs) that is consistent with the room type and layout. This is a challenging problem because the generated scene should satisfy multiple constrains, e.g., each object should lie inside the room and two objects should not occupy the same volume. To address these challenges, we propose a deep generative model that encodes these relationships as soft constraints on an attributed graph (e.g., the nodes capture attributes of room and furniture elements, such as class, pose and size, and the edges capture geometric relationships such as relative orientation). The architecture consists of a graph encoder that maps the input graph to a structured latent space, and a graph decoder that generates a furniture graph, given a latent code and the room graph. The latent space is modeled with auto-regressive priors, which facilitates the generation of highly structured scenes. We also propose an efficient training procedure that combines matching and constrained learning. Experiments on the 3D-FRONT dataset show that our method produces scenes that are diverse and are adapted to the room layout. 
\end{abstract}

\section{Introduction} \label{sec: Introduction}
The last few years have seen significant advances in image generation powered by the emergence of deep generative models such as GANs \cite{NIPS2014_5ca3e9b1} and VAEs \cite{Kingma2014AutoEncodingVB}. State-of-the-art methods are able to generate images of a single object category (e.g., faces) with amazingly realistic quality (e.g., \cite{Karras_2020_CVPR}). However, the problem of generating images of complex scenes composed of multiple objects in diverse arrangements remains a challenge. As an example, images of indoor scenes consist of room elements (floor, walls, etc.) and furniture items (table, chairs, beds, etc.) arranged in different ways depending on the room type (living room, bedroom, etc.). Moreover, room elements and furniture items should satisfy geometric constraints, e.g., each object must lie inside the room and on the floor, two objects cannot occupy the same volume, some objects tend to co-occur in particular orientations relative to the room layout. 

Recent work on indoor scene image generation \cite{Gadde_2021_ICCV} aims to address the challenge of generating complex indoor scenes by using GANs with multiple discriminators that specialize in localizing different objects within an image. By adding a ``broker'' to mediate among such discriminators,  \cite{Gadde_2021_ICCV} achieves state-of-the-art (SOTA) results on synthesizing images of living rooms. However, such SOTA image generation models are far from capturing the rich structure present in indoor scenes. For example, \cite{para2021generative} notice that these models fail to respect the relationships between scene objects and often cannot preserve certain shapes like axis-aligned polygons. We contend that addressing such complex image generation problems requires reasoning about the scene content in 3D space. 

As a stepping stone, this paper focuses on the problem of conditional generation of the scene's 3D layout, rather than a 2D image, though we can synthesize images using a renderer given the layout. Specifically, we assume we are given the room type (e.g., living room or bedroom) and the room layout (spatial arrangement of walls, windows and doors), and our goal is to generate a collection of objects (e.g., furniture items such as sofa, coffee table and chairs) that is consistent with the room type and layout. For example, a bedroom must consist of a bed, typically placed in the center of the longest wall in the room. Moreover, we expect the generator to synthesize diverse object arrangements for the same room.
This problem of conditional 3D layout generation is important in applications such as room decoration, where the goal is to produce diverse decors for a given room.

Recent work \cite{armeni20193d,wang2018deep,keshavarzi2020scenegen} aims to address 
this problem
using supervision in the form of scene hierarchies or relational graphs. However, the contextual space of possible arrangements objects in a room is simply too large to be modeled using hand-crafted heuristics or hierarchies. This has led to recent efforts on training networks directly from data using autoregressive models based on CNNs \cite{Ritchie2019FastAF} or Transformers \cite{wang2020sceneformer,Paschalidou2021NEURIPS}. These models, while being adept at generating indoor scenes, lack the advantages of a learnt latent space as in Variational Autoencoders (VAEs). The VAE latent space is often a good representation of data which can bootstrap several downstream applications. In this work, we show one such application of furniture recommendation given a floorplan by retrieving the most ``appropriate" furnished room from a database curated by human designers and adapting it to the new floorplan. The presence of a learnt latent space allows us to naturally define a notion of ``appropriateness" which is otherwise a non-trivial problem (more details in \S\ref{sec. experiments}). Moreover, users can often traverse the latent space to manipulate generated samples allowing more controlled generations \cite{higgins2016beta,kumar2017variational} (see Figure \ref{Figure: Latent space traversals}). Such manipulations are not easy to implement in autoregressive models. Unfortunately, we found in our experiments that existing graph-based VAE architectures are insufficient for indoor scene generation. This observation is echoed by Para \etal \cite{para2021generative} in their work on 2D layout generation, and they conjecture that current VAE architectures struggle with the discrete nature of graphs and layouts.

To remedy this, we propose a graph-based VAE model for synthesis of 3D indoor scenes conditioned on the room type and layout (floorplan). We represent both the room and furniture layouts with an attributed graph. We then present a scene generative model consisting of a graph encoder that maps the input graph to a latent space, and a graph decoder that generates a furniture graph, given a latent code and the room graph. Our model considerably reduces the performance gap between VAEs and state-of-the-art autoregressive models \cite{Paschalidou2021NEURIPS} for indoor scene synthesis. Specifically, we make the following contributions: 
\begin{enumerate}[leftmargin=*,noitemsep]
\item \emph{A structured auto-regressive prior for graphs}: This is our main contribution. Contemporary graph-VAE architectures typically encode the graph into a single latent vector and then use a multi-layered perceptron (MLP) to decode it back to a graph \cite{simonovsky2018graphvae,Kipf2016VariationalGA}. In contrast, we propose to have a separate latent code for each furniture item. This has been previously explored in \cite{luo2020end} where the authors assume an i.i.d. Gaussian prior over the latents. This limits performance since the graph decoder struggles to learn complex relationship between different furniture nodes from i.i.d latent codes as input.\footnote{This is corroborated by our experiments in \S\ref{sec. experiments} where we consider this architecture as Baseline B1.} Instead, we propose a novel autoregressive prior based on linear Gaussian models which allows the model to learn a dependency structure between the different latent variables corresponding to different furniture items in the scene. We also propose an efficient way to compute the KL divergence term in the VAE objective which requires a  matching procedure since there is no canonical ordering of graph nodes. 

\item \emph{Learning graph-VAEs under constraints:} To facilitate learning, we use simple intuitive constraints like limiting the relative distances between furniture items, such as a chair and a table. These can be easily computed from training data. We then train our VAE model under these constraints utilizing a recently introduced constrained learning framework \cite{chamon2020probably}. 

\item \emph{Experimental evaluation}: Through extensive experiments we show that our proposed model achieves considerable improvement over baseline VAE architectures bringing the performance of latent-variable models at scene synthesis closer to what can be achieved using autoregressive models. Moreover, we show a unique application of our latent-variable model which is not possible with autoregressive models. Finally, we show how one can edit the generated scenes post-hoc by traversing the latent space.
\end{enumerate}

\section{Related Work} \label{sec:related}


\textbf{Graph-based inference:} Graphical representation of scenes and graph-based inference has been extensively studied in the past. Early works \cite{fisher2012example,fisher2011characterizing,yeh2012synthesizing,jiang2018configurable,qi2018human} employed ``shallow" methods like hand-crafted graph kernels or probabilistic graphical models to learn the furniture arrangements. 
Recent works leverage deep generative models to learn good scene representations directly from spatial data. The community has explored avenues for combining graphs with VAEs to synthesize 3D scenes \cite{10.1145/3303766,zhang2020deep,luo2020end}. However, all these methods rely on strong heuristics on defining object relations. For instance, \cite{luo2020end} relies on user-defined scene-graphs as input, \cite{10.1145/3303766} requires hand-crafted hierarchies, and \cite{Purkait2020SGVAESG} uses heuristics to extract context-free grammars from data which are then used to train a grammar-VAE \cite{kusner2017grammar}. Our proposed VAE is different from all these methods in that we do not use any such strong heuristics on object relations, but only a check of association between a furniture item and a room element using a distance measure. On the other hand, Wang \etal \cite{wang2019planit} uses graphs for high-level planning of the furniture layout of the room in a 2-stage approach where they train a generator to synthesize scene graphs followed by a CNN to propose consistent furniture poses. Their model has no latent variables and is slower due to the $2$-stage process. 

\textbf{Autoregressive Scene Generation.} Recent successful models for indoor scene synthesis are all autoregressive in nature \cite{wang2018deep,Ritchie2019FastAF,wang2020sceneformer,Paschalidou2021NEURIPS}. Wang et al. \cite{wang2020sceneformer} introduced an autoregressive scene generation pipeline, Sceneformer, using multiple transformers \cite{NIPS2017_3f5ee243} which predict objects' category, location and size separately.  Concurrently, FastSynth \cite{Ritchie2019FastAF} introduced a similar pipeline, where the authors train separate CNNs based on a top-down representation of the scene to sequentially insert objects into the scene. Their method however requires auxilary supervision in the form of depth and semantic segmentation maps. Another recent transformer-based autoregressive approach was proposed in \cite{Paschalidou2021NEURIPS}, which replaces the multiple trained models of past works with a single unified model. Unlike these models, we learn an end-to-end latent variable model to generate 3D indoor scenes trained from spatial data.  

\textbf{Expressive latent distributions for VAEs.} There has been extensive work into designing expressive distributions for VAEs. For example, \cite{klushyn2019learning} proposes a hierarchial prior, \cite{rezende2015variational} uses normalizing flows to model a more expressive posterior distribution over latents, \cite{chen2016variational} uses an autoregressive prior, and \cite{tomczak2018vae} advocates the use of a mixture distribution for the prior based on the posterior distribution of the encoder. However, learning expressive distributions for latent spaces which are expressed as graphs is challenging due to the absence of any canonical ordering between the different nodes of a graph. This makes computing the required KL divergence term in the ELBO notoriously difficult. In this work we propose to model the latent space as an autoregressive linear Gaussian model which allows us to formulate the ordering as a Quadratic Assignment problem for which we also propose an efficient approximation.

\section{Our Approach} \label{sec:method}
  \begin{figure}[ht]
 	\centering
 	\includegraphics[width=1.0\linewidth]{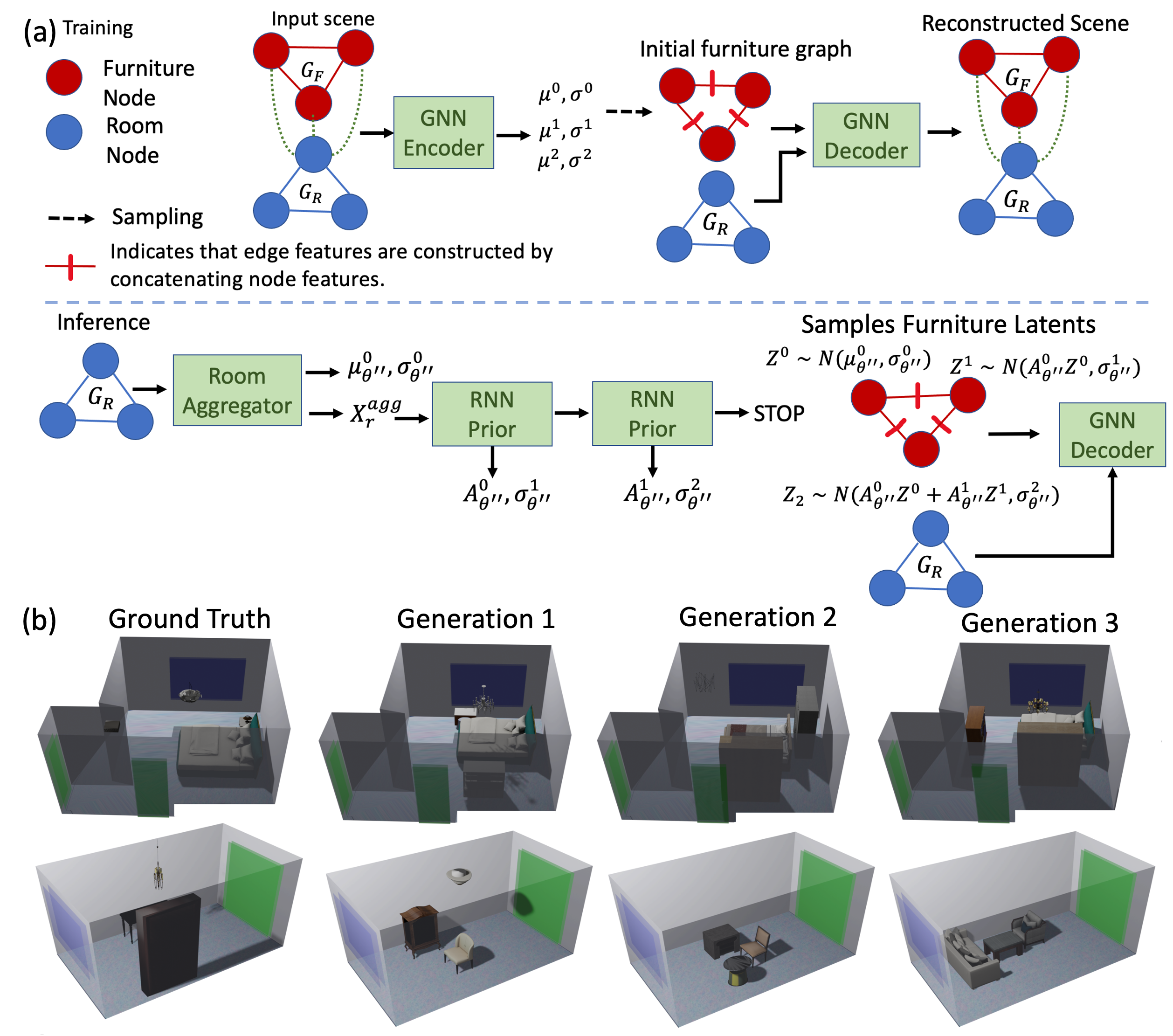}
 	\caption{\textit{\textbf{(a) Overview of the proposed model.} The input scene graph includes the furniture and room sub-graphs and is complete (we omit depicting certain edges to prevent clutter). The Encoder predicts a mean and variance per furniture node, which are then used to sample latents for proposing furniture nodes by the decoder. During inference we use the proposed autoregressive prior to generate the latents, which are then subsequently processed by the decoder for scene synthesis; \textit{\textbf{(b) Generated scenes.} Sample generations for our model for a bedroom (row 1) and a library (row 2). Green rectangles indicate doors while blue rectangles indicate windows}}}
 	\label{fig: proposed architecture}
 \end{figure}
This section describes the proposed model. First, we describe how we represent the 3D scene layout with an attributed graph. Next, we describe our architecture. The VAE encoder is a Graph Neural Network (GNN) that processes the graph and produces latent variables which are then passed to another GNN which serves as the VAE decoder. Then, we describe how we parameterize the prior on the latent space using an autoregressive model which is learnt. Finally, we describe the proposed training methodology, which includes some constraints for faster convergence. A schematic depiction of our approach is given in Figure \ref{fig: proposed architecture}a with more details in the Appendix.

\subsection{Indoor scene representation as a graph}
\label{subsection: indoor scene representation as a graph}
We represent an indoor scene as an attributed graph $G=(V, E, X)$. Here the nodes $V$ denote room layout components (floor, wall, windows) and furniture items (sofa, chair, bed), the edges $E \subseteq V \times V$ denote relationships between the nodes (e.g., relative orientation of sofa to wall), and the attributes $X$ denote features associated with the nodes and edges (e.g., location of furniture items or relative orientation between bed and wall). The graph has two node types (room nodes $V_R$, furniture nodes $V_F$), three edge types (room-room edges $E_{RR}$, room-furniture edges $E_{RF}$ and furniture-furniture edges $E_{FF}$), and five attribute types corresponding to these node and edge types ($X_R$, $X_F$, $X_{RR}$,  $X_{RF}$ and $X_{FF}$). 
 In this work, we consider the graph as complete, that is, $E_{RF} = V_R \times V_F, E_{FF} = V_F \times V_F$ and $E_{RR} = V_R \times V_R$. 
 
We will identify two main subgraphs of $G$. The room layout graph $G_R = (V_R,E_R,{X_R,X_{RR}})$ consists of $n_R := |V_R|$ nodes (or room elements) and $e_R := |E_R|$ edges, where the node attributes $X_R \in \sR^{n_R \times d_R}$ denote the class, location, orientation and size of the room element, and the edge attributes $X_{RR} \in \sR^{e_R \times d_{RR}}$ encode geometric or functional relationships between two room elements (relative location, relative orientation etc.). The room type $T$ is also encoded as a categorical feature into $X_R$. Similarly, the furniture layout graph $G_F=(V_F,E_F,{X_F,X_{FF}})$ consists of $n_F := |V_F|$ nodes (or furniture items) and $e_F=|E_F|$ edges, where the node attributes $X_F \in \sR^{n_F \times d_F}$ denote the class of the furniture item, its location, orientation, size, and its 3D shape descriptor, and its edge attributes $X_{FF}  \in \sR^{e_F \times d_{FF}}$ encode geometric or functional relationships between two furniture items. We obtain the shape descriptors of each furniture item by processinga a 3D point cloud of the item through PointNet~\cite{qi2017pointnet} pretrained on the Stanford ShapeNet dataset \cite{shapenet2015}. 
 
 \subsection{Proposed Generative Model}
 \label{subsection: proposed generative model}
We would like to design and learn a probabilistic model $p(G_F \mid G_R,T,n_F)$ that generates a furniture layout $G_F$ given the room layout $G_R$, type $T$ (say, bedroom or library room) and number of furniture items $n_F$ to place in the room. 
We assume there exists latent $Z$ such that ~ $p(G_F \mid  G_R, T, n_F)~ =$
\begin{equation}
  \! \!\! \int \!\! p(G_F \!\! \mid \!\! Z,  n_F, G_R, T)   p(Z \!\! \mid \!\! n_F, G_R, T)dZ .
	\label{eq: generative model}
\end{equation}
Our proposed model consists of three main ingredients: 

\begin{enumerate}[leftmargin=*,noitemsep] 
	\item An encoder, $q_{\phi}(Z \mid G, T, n_F)$, which maps both room and furniture layouts ($G_R$ and $G_F$ resp.) as well as room type and number of furniture items to a latent variable $Z$, which captures the diversity of room-aware  furniture layouts. The parameters of the encoder are denoted as $\phi$. 
	\item A decoder, $p_{\theta'}(G_F \mid  Z,  n_F, G_R, T)$, which maps the number of furniture items, the latent variable as well as the room layout and type to a furniture layout. The parameters of the decoder are denoted as $\theta'$. 
	\item A prior model $p_{\theta''}(Z \mid  n_F, G_R, T)$. The difference between the prior model and the encoder is that the prior model only considers the room layout $G_R$ and not $G$. The parameters of the prior model are denoted as $\theta''$. 
\end{enumerate}
The encoder, decoder and prior model are parameterized with GNNs which we will describe in the next subsection. 

Given a training set, consisting of indoor scenes in the form of attributed graphs $\gG = \{G_1, G_2, ..., G_n\}$, we learn the parameters $\{\theta', \theta'', \phi\}$ by optimizing the following empirical average over the training set:
\begin{equation}
	\begin{aligned}
		\gL(\theta', \theta'', \phi) = & \frac{1}{n} \sum_{i=1}^n \mathscr{L}(G_i, \theta', \theta'', \phi),
	\end{aligned}
	\label{eq: overall ELBO}
\end{equation}
where, for any $G \in \gG$, $L(G, \theta', \theta'', \phi)$ denotes the Evidence Lower Bound (ELBO) defined as:
\begin{equation}
	\begin{aligned}
		&\mathscr{L}(G, \theta', \theta'', \phi) \\
		&=  \mathbb{E}_{Z \sim q_{\phi}(Z \mid G, T, n_F)} [\log p_{\theta'}(G_F \mid  Z,  n_F, G_R, T)] \\
		& - KL(q_{\phi}(Z \mid G, T, n_F) \mid \mid p_{\theta''}(Z \mid  n_F, G_R, T) ).
	\end{aligned}
\label{eq: ELBO}
\end{equation}
As mentioned in \S\ref{sec: Introduction}, we optimize \eqref{eq: overall ELBO} under constraints to speed up convergence, as we will describe in subsection \ref{section: learning under constraints}.

\textbf{Graph Encoder.} The encoder models the approximate posterior of a latent variable $Z$ given $(G,T,n_F)$. We assume that the distribution of $Z$, $q_{\phi}(Z \mid G,T,n_F)$, is Gaussian with mean $\mu_{\phi}(G,T,n_F)$ and a diagonal covariance matrix with diagonal entries $\sigma_{\phi}(G,T,n_F)$. The  distribution parameters $(\mu_{\phi}, \sigma_{\phi})$ are modeled as the output of an attention-based message passing graph neural network (MP-GNN) with weights $\phi$.  The design of the MP-GNN is inspired by \cite{mavroudi2020representation}, where each layer $l=1,\ldots,L$ of the MP-GNN maps a graph $G^{l-1}$ to another graph $G^l$ by updating the graph’s node and edge features. Specifically, let $h_i^l$ and $h_{ij}^l$  denote the features of node $i$ and edge $(i,j)$ of graph $G^l$, respectively. Let the input to the network be the graph $G^0=G$, so that $h_i^0$ and $h_{ij}^0$ denote the node features (rows of $X_R$ and $X_F$) and edge features (rows of $X_{RR}$, $X_{FF}$ and $X_{RF}$), respectively. At each iteration of node and edge refinement, the MP-GNN: (1) adapts the scalar edge weights by using an attention mechanism; (2) updates the edge attributes depending on the edge type, the attention-based edge weights, the attributes of the connected nodes and the previous edge attribute; and (3) updates the node attribute by aggregating attributes from incoming edges.

After $L$ layers of refinement, we obtain a graph $G^L$, whose node features are mapped via a linear layer with weights $W_{\mu}, W_{\sigma}$ to obtain the parameters of the Gaussian model as $\mu_\phi^i(G, T) = W_{\mu} h_i^L, \; \sigma^i_\phi(G,T) = \exp(W_{\sigma} h_i^L)$
where $i=1,\dots, n_F$.
Note that there is a different Gaussian for each node of the graph. Therefore, the output of the encoder will be two matrices $\mu_\phi(G,T,S) \in\mathbb{R}^{n_F \times d_F}$  and $\sigma_\phi(G,T,S) \in\mathbb{R}^{n_F \times d_F}$ corresponding to the mean and standard deviation vectors of the latent variable matrix $Z\in\mathbb{R}^{n_F \times d_F}$. 

 \textbf{Graph Decoder.} The decoder maps $(n_F, Z)$ and the room layout, type $(G_R, T)$ to a desired furniture layout via the distribution $p_{\theta'}(G_F \mid  Z,  n_F, G_R, T)$. The generative process proceeds as follows,

An initial fully connected furniture graph $G_F^0$ is instantiated.  Each node of $G_F^0$ is associated with a feature of dimension $d_F$ corresponding to one of the rows of $Z$.  Each edge $(i,j)$ of $G_F^0$  of type $\epsilon \in\{RF,FF\}$ is associated with a feature $Z_{ij} = (Z_i,Z_j)$ as the concatenation of the node features. As a result, we obtain an initial graph $G_0$ that includes both the initial furniture graph $G_F^0$ as well as the given room graph $G_R$ as subgraphs. The initial graph $G_0$ is passed  to a MP-GNN, which follows the same operations as the encoder MP-GNN.

The output of the MP-GNN is the furniture subgraph $G_F^L$ of the final graph $G^L$. Each furniture node of $G^L$ is then individually processed through a MLP to produce parameters for the furniture layout graph distribution $p_{\theta'}(G_F \mid Z, n_F, G_R, T)$,  which can be factorized in the following way:
	\begin{align}
		&p_{\theta'} (G_F \mid Z, n_F, G_R, T) \\=& \prod_{i=1}^{n_F}p_{\theta'}(shape_i \mid Z, G_R, T) p_{\theta'}(orien_i \mid Z, G_R, T) \nonumber \\
		&p_{\theta'}(loc_i \mid Z, G_R, T)p_{\theta'}(size_i \mid shape_i) p_{\theta'}(cat_i \mid shape_i). \nonumber
	\end{align}
	More specifically, we assume that given latent $Z$, room layout $G_R$ and type $T$, the furniture features are independent of each other (a standard assumption in the VAE literature). For a furniture, we further assume that shape, orientation and location features are independent given $Z, G_F$ and $T$. However, since the PointNet shape features implicitly capture the configuration of the furniture item in 3D space we condition the size and category distributions on the shape feature. 

We parameterize the shape and location features as a normal distribution, the size feature as a lognormal distribution whose support is restricted to be positive valued, category and orientation features as categorical distribtions.  Since both the Encoder and Decoder graphs have $n_F$ nodes that are in one-to-one correspondence, we can define our reconstruction loss (first term in \eqref{eq: ELBO}) by simply comparing their node features without the need for an explicit matching procedure.

\begin{figure*}[h]
	\centering
	\includegraphics[width=0.85\linewidth]{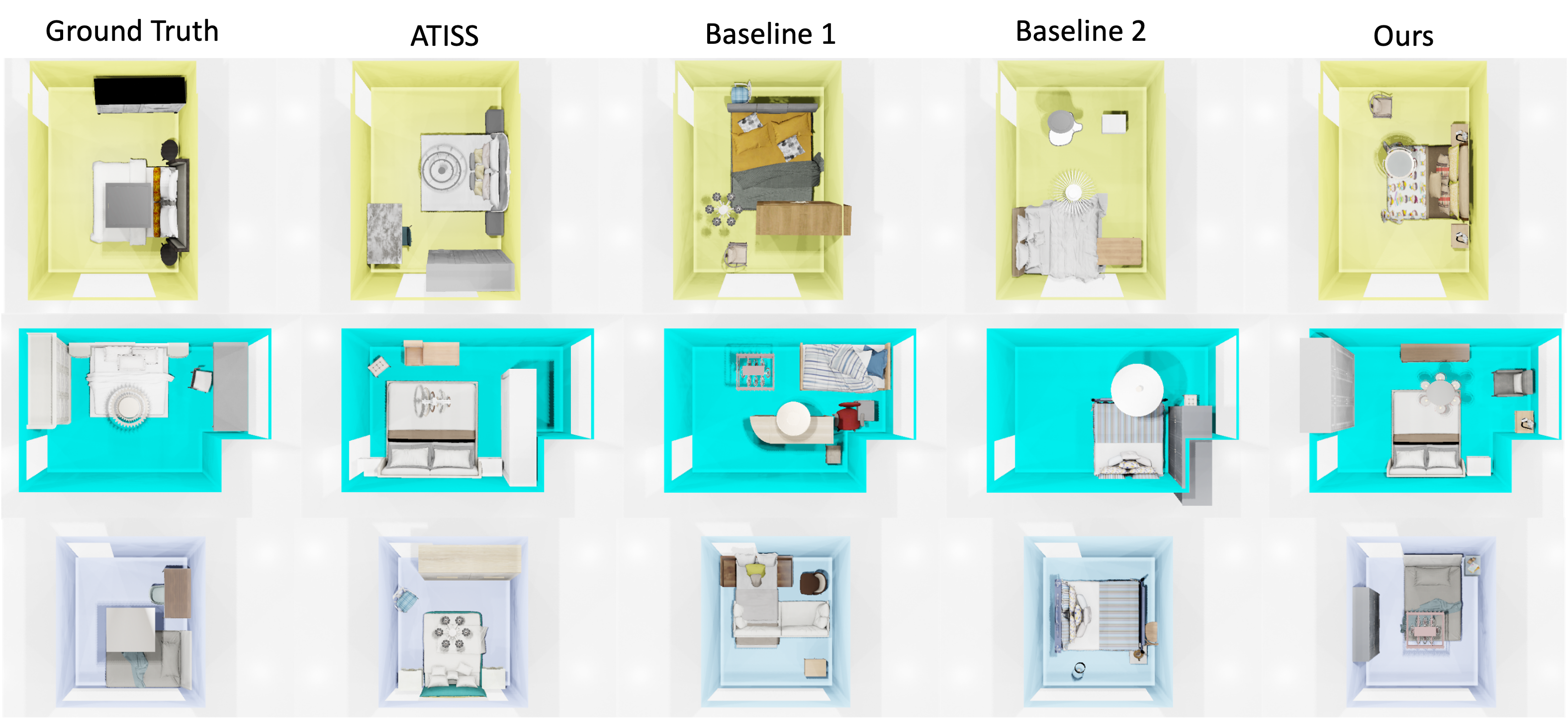} 
	\caption{\textit{Qualitative comparison of our method with ATISS and baselines. Windows and doors and indicated by white rectangles}}
	\label{fig: qualitative comparison}
	\vspace{-4mm}
\end{figure*}

\textbf{Graph Prior.} Recall our latent space $Z$ is modelled such that there is a latent variable corresponding to each furniture node in the graph.  Many popular graph VAE models assume an i.i.d. normal prior for each node \cite{kipf2016variational,luo2020end}. However, such a model is restrictive for our purposes. MP-GNNs achieve permutation equivariance by sharing the weight matrices across every node in the graph. When the graph is complete, as is the case here, the marginal distribution of every output node after $L$ GNN layers will be identical if they are initialized as i.i.d. Gaussian at the input layer. Since, the output nodes of the decoder GNN after $L$ layers correspond to different furniture features, having identical marginals is detrimental. This claim is supported by experiments (Figure \ref{fig: qualitative comparison}) where the i.i.d. prior baseline models struggle to learn proper furniture placements. To remedy this, we propose to parameterize prior distribution as an autoregressive model based on linear gaussian models \cite{bishop2006pattern}. More specifically, 

\begin{align}
	p(Z^0 \mid G_R, T)\! &=\! \gN(\mu_{\theta''}(G_R, T), \sigma^0_{\theta''}(G_R, T)), \label{eq: autoregressive model} \\
	p(Z^i | Z^{k < i}, G_R, T) \! &= \! \gN(\sum_{k < i}A^k_{\theta''}(G_R, T) Z^k, \sigma^i_{\theta''}(G_R, T)). \nonumber \\
\end{align}

where $Z^i $ refers to the latent corresponding to the $i^{th}$ furniture node. Thus, the $i^{th}$ furniture node latent is given by a Gaussian whose mean is a linear function of all the latents $k < i$. Such a structure ensures that all the latent variables are jointly Gaussian.  This allows us to analytically compute the KL divergence term and thus was favoured over more expressive probabilistic models which would introduce more stochasticity in the objective due to the need of estimating the KL divergence term via sampling. We implement \eqref{eq: autoregressive model} (see also Figure 1) with two networks:
\begin{itemize}[leftmargin=*,noitemsep] 
	\item \textbf{Room Aggregator for $\boldsymbol {p(Z^0 \mid G_R, T)}$:} The room aggregator is an MP-GNN with the same architecture as the Graph Encoder. except that the input to the network is just $(G_R, T)$ with the node and edge features initialized to $X_R$ and $X_{RF}$, respectively. After $L$ GNN layers, all the room node features are aggregated by a mean pooling operation to obtain a global representation of the room layout plan $X_R^{agg}$. This $X_R^{agg}$ is then passed through an MLP to compute $\mu_{\theta''}(G_R, T)$ and $\sigma^0_{\theta''}(G_R, T)$.
	\item \textbf{RNN Prior for $\boldsymbol {p(Z^i | Z^{k < i}, G_R, T)}$:} We use a recurrent neural network to predict the matrix $A^k_{\theta''}(G_R, T)$ and the variance $\sigma^i_{\theta''}(G_R, T)$ at each node index. The RNN is initialized with $X_R^{agg}$. We need additional constraints on each $A^k_{\theta''}(G_R, T)$ to prevent the dynamics model in \eqref{eq: autoregressive model} from diverging to infinity. This is typically done by controlling the spectral radius or its proxy, the spectral norm \cite{lacy2003subspace}, of the matrices $\{A^k_{\theta''}(G_R, T): k \in [1, 2, ..., n_F]\}$. Thus, the  predicted matrix is taken to be $\frac{A^k_{\theta''}(G_R, T)}{||A^k_{\theta''}(G_R, T)||_{2}}$, where $||A||_2$  is the spectral norm of some matrix $A$.
\end{itemize}

\subsection{Computing the KL term via matching and constrained learning}
 \label{section: learning under constraints}
Note that the proposed autoregressive prior could in principle be reexpressed as a more traditional i.i.d.~Gaussian prior, which is then passed through an additional non-equivariant transformation layer that can be absorbed into the decoder.  But a significant difference emerges in practice when facing the key challenge of incorporating this non-equivariant factor into subsequent model training, given that there is no longer a canonical ordering between the different nodes in a graph.   When the proposed autoregressive prior formulation is adopted, such an ordering is only required to evaluate $p_{\theta''}(Z \mid G_R, T, n_F)$ for any  $Z$ sampled from the posterior $q_{\phi}(Z \mid G, T, n_F)$ to compute the KL divergence term in the ELBO \eqref{eq: ELBO}.  However, by design this term can be expressed analytically, and as we will soon demonstrate, an efficient compensatory permutation can be efficiently computed.  In contrast, with an alternative autoregressive decoder formulation, the search for an appropriate ordering is instead needed for computing the VAE reconstruction term (i.e., evaluating the decoder $p_{\theta'} (G_F \mid Z, n_F, G_R, T)$ for any $Z$ sampled from the posterior), and hence becomes entangled with the non-analytic stochastic sampling required for obtaining approximate reconstructions.

\textbf{Computing the KL divergence term.} Let $Z = \{Z^1, Z^2, ..., Z^{n_F}\}$ be the set of latent variables correponding to $n_F$ furniture items to be placed in the room. Let $\pi$ denote the ordering among these variables. Given $\pi$, the likelihood of observing $Z$ under our proposed prior is defined as 
\begin{equation}
	p_{\theta''}(Z \mid G_R, \pi, n_F, T) = \prod_{i=1}^{n_F} p_{\theta''}(Z^{\pi(i)} \mid Z^{\pi(j < i)}, G_R, T),
	\label{eq. graph prior based on ordering}
\end{equation}
However, for $Z \sim  q_{\phi}(Z \mid G, T, n_F)$ (the approximate posterior) we do not know this ordering $\pi$. Thus, given $Z$, we define the optimal order $\pi^*$ to be 

\begin{equation}
	\pi^*\! =\! \argmin_\pi KL( q_{\phi}(Z \!\mid\! G, T, n_F)\! \mid\mid \! p_{\theta''}(Z\! \mid\! G_R, \pi, n_F, T)).
	\label{eq: optimal ordering}
\end{equation}
Recall $q_{\phi}(Z \mid G, T, n_F)  = \prod_{i=1}^{n_F}\gN(Z^i; \mu_{\phi}^i(G), \sigma_{\phi}^i(G))$
Since both the prior and posterior are jointly Gaussian, computing \eqref{eq: optimal ordering} reduces to solving a Quadratic Assignment Problem (QAP).  For simplicity let us denote the distributions as
\begin{equation}
	\begin{aligned}
		q_{\phi}(Z \mid G, T, n_F) &= \gN(\mu_0, \Sigma_0) \\ p_{\theta''}(Z \mid G_R, \pi, n_F, T) &= \gN(\tilde{\pi}\mu_1,\tilde{\pi}\Sigma_1\tilde{\pi}^T),
	\end{aligned}	
\end{equation}
where $\tilde{\pi} = \pi \otimes I_{d_F \times d_F}$. Note $\pi \in \sR^{n_F \times n_F}$ and the kronecker product comes from the fact that we are only allowed to permute blocks of $\mu_1$ and $\Sigma_1$, of size $d_F$ and $d_F \times d_F$ respectively, where $d_F$ is the dimension of the latent variable used for each furniture node. In other words, we can only permute latent vectors corresponding to furniture nodes as a whole and not the intra dimensions of $Z$ within any furniture node. Here $\mu_0, \mu_1 \in \sR^{n_Fd_F}$. Similarly, $\Sigma_0, \Sigma_1 \in \sR^{n_Fd_F \times n_Fd_F}$. Note that due to the autoregressive structure $\Sigma_1$ will not be a diagonal matrix. 

We show in the Appendix (\S\ref{subsection: Derivation for KL term}) that \eqref{eq: optimal ordering} is equivalent to
\begin{equation}
 \min_{\pi} Tr\left(\Sigma_1^{-1} \tilde{\pi}^T \left[ \Sigma_0 + \mu_0\mu_0^T \right]\tilde{\pi}\right) - 2Tr\left(\tilde{\pi}^T \mu_0\mu_1^T\Sigma_1^{-1}\right).
  \label{eq. KL div. analytical form}
\end{equation}
Notice that \eqref{eq. KL div. analytical form} is a Quadratic Assignment Problem (QAP) which is known to be NP-Hard. For this we propose to use a fast approximation algorithm, called FAQ, introduced in \cite{vogelstein2015fast}. FAQ first relaxes the optimization problem from set of permutation matrices to the set of all doubly stochastic matrices. It then iteratively proceeds by solving linearizations of the objective \eqref{eq. KL div. analytical form} using the Franke-Wolfe method. These linearizations reduce the QAP to just a linear assignment problem (LAP), which can be efficiently solved by the Hungarian algorithm. After Franke-Wolfe terminates, we project the doubly stochastic solution back to the set of permutations by solving another LAP. FAQ has a runtime complexity that is cubic in the number of nodes per iteration which is faster than the quartic complexity of the matching procedure used in \cite{simonovsky2018graphvae}. More details in Appendix \S\ref{subsection: Solving using the FAQ approximation}. 

We need to compute the optimal $\pi^*$ for each graph in a mini-batch, and then compute the KL divergence term in \eqref{eq: ELBO} analytically given this ordering. We observe no significant gains in performance in running the FAQ algorithm more than $1$ step per graph, which further speeds our method. 

 \textbf{Learning under constraints:} To facilitate faster convergence and also ensure fidelity of the learned solution, we enforce certain constraints on the reconstructed room. These constraints are derived from training data and do not require external annotations. Given input furniture graph $G_F$ to the encoder and the reconstructed graph $\tilde{G}_F$ by the decoder, we enforce the relative positions of predicted furnitures in $\tilde{G}_F$ to be ``close" to the ground truth relative positions in $G_F$. Similarly, we apply constraints on the relative position of the predicted furniture items with the room walls, windows and doors. Finally we apply a constraint penalizing the relative orientations between different furniture items from being too ``far" away from the relative orientations in $G$. We explain these constraints more clearly in Appendix \ref{appendix: learning under constraints}.

Having described all the ingredients in our model, we finally present the complete optimization objective as 

\begin{equation}
	\begin{aligned}
		\max_{\theta', \theta'', \phi} \gL(\theta', \theta'', \phi) \quad s.t. \qquad \frac{1}{n} \sum_{i=1}^n  \textrm{ Constr}(G_i) \leq \epsilon. \\
	\end{aligned}
\end{equation}
Here $\epsilon$ is a user-defined hyperparameter that determines the strictness of enforcing these constraints. $\gL(\theta', \theta'', \phi) $ is as defined in \eqref{eq: overall ELBO} and $i$ is in iterator over the scene graphs in the training set. We employ the learning under constraints framework introduced by \cite{chamon2020probably} which results in a primal-dual saddle point optimization problem. We give exact algorithmic details in the Appendix (Algorithm \ref{algorithm: constraints}).

\subsection{Inference}
For inference, we start with a room layout graph $G_R$ and type $T$ along with the number of furnitures to be placed in the room $n_F$. We then use the learnt autoregressive prior to sample $n_F$ latent variables recursively. This latent $Z$ along with $(n_F, G_R, T)$ is processed by the graph decoder to generate the funiture layout subgraph $G_F$.  For scene rendering, we use the predicted shape descriptor for each furniture item and perform a nearest neighbour lookup using the $\ell_2$ distance against a database of 3D furniture mesh objects indexed by there respective PointNet feature. We then use the predicted size and orientation to place furnitures in the scene. Note the inclusion of shape descriptors allows the model to reason about different appearance of furniture items that is more room aware as opposed to retrieval based on just furniture category \& size as in \cite{wang2020sceneformer,Paschalidou2021NEURIPS}.  

\section{Experiments} \label{sec:results}
\label{sec. experiments}
 \begin{figure}
	\centering
	\includegraphics[width=1.0\linewidth]{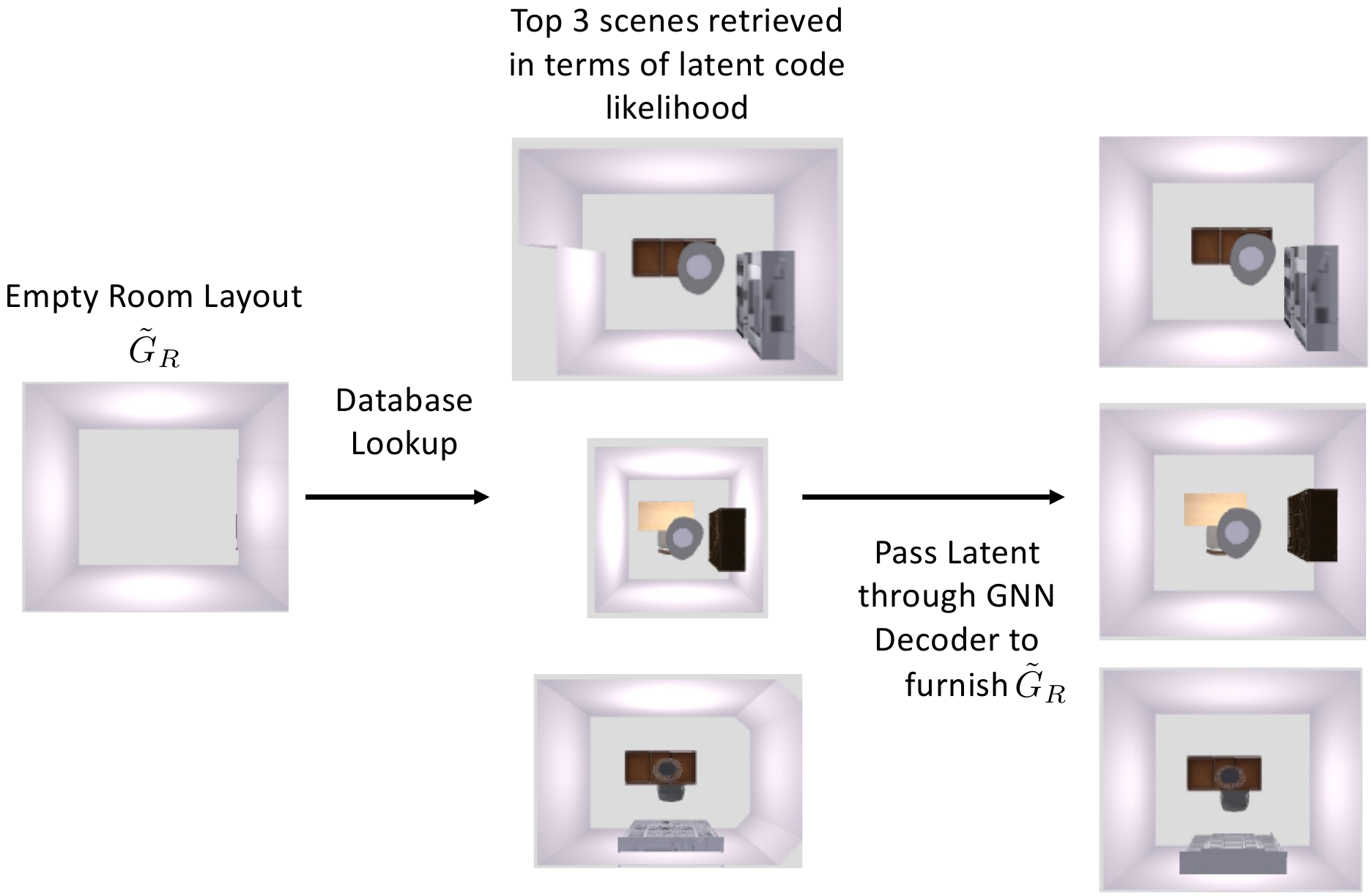}
	\caption{\textit{Manipulating Latent Space for Design Recommendations from Database.}}
	\label{fig: database retrieval}
	\vspace{-4mm}
\end{figure}

In this section, we evaluate the effectiveness of the proposed approach qualitatively and quantitatively. More details and figures provided in the Appendix. 

\textbf{Dataset:} We use the 3D-FRONT dataset \cite{fu20213d} for all our experiments. The dataset consists of roughly $14k$ rooms, furnished with $3$D mesh furniture items. We consider 4 room types - bedroom, living room, library \& dining room. 

\textbf{Training Protocols:} All models were trained on a $80:20$ train-test split. Since we use PointNet shape descriptors, we only predict 7 ``super-categories"\footnote{Cabinet/Shelf, Bed, Chair, Table, Sofa, Pier/Stool, Lighting.} of furniture items.\footnote{This is contrary to prior work \cite{Paschalidou2021NEURIPS} which explicitly models the 34 fine-grained categories in 3D-FRONT. We made this design choice since we use shape descriptors for each furniture item which has the fine-grained label information encoded in them. The $7$ super-categories are only used to provide high-level supervision to distinguish between furnitures which may have similar shapes, for example, a chair and a sofa-chair. During synthesis, our model predicts the shape descriptors, which is used for rendering, and thus can generate furniture's belonging to all the 34 categories.} Since furniture is mostly axis-aligned, we discretized the orientation into four categories $[0^{\circ}, 90^{\circ}, 180^{\circ}, 270^{\circ}]$. We also rotate the room by multiples of $90^{\circ}$ as data augmentations. We train all models using the ADAM optimizer for $1500$ epochs with a batch size of $128$. 

\textbf{Baselines:} We compare our VAE model with baselines which have the same Encoder-Decoder architecture but employ the commonly used i.i.d. prior. We compare against two variants (inspired from \cite{luo2020end}), $(i)$ \textbf{Standard Prior (B1):} Each $Z^i \sim \gN(0, I)$ for $i \in [1, ..., n_F]$, $(ii)$ \textbf{Non-autoregressive Learnt Prior (B2):} Each $Z^i \sim \gN(\mu(G_R), \sigma(G_R))$. The mean and variance parameters of this distribution are learnt by an MP-GNN which processes just the room subgraph. 


\begin{table}

\caption{\textit{Quantitative comparison of our method with other models and baselines. Scene classification accuracy closer to 0.5 is better}}
    \centering
    \resizebox{1.0\linewidth}{!}{
    \begin{tabular}{l|cccc|cccc}
        \toprule
        \multicolumn{1}{c}{\,} & \multicolumn{4}{c}{Category KL Divergence ($\downarrow$)}& \multicolumn{4}{c}{Scene Classification Accuracy}\\
        \toprule
        \multicolumn{1}{c}{\,} & Bedroom & Living & Dining & Library & Bedroom & Living & Dining & Library\\
        \midrule
        ATISS & \textbf{0.01} & \textbf{0.00} & \textbf{0.01} & \textbf{0.01} & \textbf{0.75} & \textbf{0.82} & \textbf{0.75} & 0.86 \\
        Baseline B1 & 0.02 & 0.02 &0.05 & 0.09 & 0.94 & 0.93 & 0.90 & 0.91 \\
        Baseline B2 & 0.02 & 0.02 & 0.05 & 0.07 & 0.93 & 0.91 & 0.85 & 0.90 \\
        Ours & \textbf{0.01} & 0.01 & 0.02 & 0.02 & 0.83 & 0.88 & 0.78 & \textbf{0.82} \\
        \bottomrule
    \end{tabular}}
    \label{Table: quantitative metrics}
\end{table}

\textbf{Scene Generation:} In Figure \ref{fig: proposed architecture}b, we illustrate the effectiveness of the proposed method at generating diverse furniture recommendations given a room layout. Notice how our model generates diverse furniture arrangements and 3D appearances e.g. sampled beds, cabinets and ceiling lights look distinct in row 1. In the library (row 2), our model proposes diverse arrangements such as a simple study room or a lounge with a couch and chair. In Figure \ref{fig: qualitative comparison} we qualitatively compare our generations for different rooms with ATISS and VAE baselines. The figure shows that our baseline models have inconsistent overlapping furniture placements. In contrast, our model generates plausible furniture arrangements, bridging the performance gap between latent-variable models and autoregressive ones (ATISS).


In Table \ref{Table: quantitative metrics}, we provide quantitative metrics comparing our model with baseline VAEs \& ATISS. The Category KL divergence measures how well the model captures the frequency of categories present in an indoor scene using all 34 fine-grained furniture labels compared to the ground truth. To obtain a fine-grained label for a generated furniture, we look up the corresponding category of the nearest-neighbour furniture retrieved using the predicted shape features. On this metric, our method is competitive with ATISS showing that even though we explicitly only model ``super-categories", our model is able to capture well the fine-grained object label frequencies of the test set via the shape descriptors.
Scene classification accuracy (close to $0.5$ is better) tests the ability of a network to distinguish between real and synthetic scenes. Different from prior work \cite{Paschalidou2021NEURIPS} we evaluate this metric by training a GNN directly on the synthesized 3D scene graphs v/s ground truth graphs, instead of classifying the rendered 2D top-down orthographic projections. This is because at the image level, CNNs are not able to explicitly reason about 3D spatial relationships compared to a GNN. Moreover, classifying 2D projections is sensitive to the rendering used and thus will not be comparable across different works. On this metric, our method performs significantly better than baselines (7-10\% improvement) and is competitive with ATISS.

\textbf{Generation Time:} The average run-time for scene synthesis for our method is \textbf{130.26ms} v/s 148.51ms for ATISS (measured on NVIDIA GeForce GTX 2080 Ti machine). 

\textbf{Do the constraints help?} Employing constraints facilitates faster training. We illustrate this in the Appendix (Figure \ref{fig: constraints}). Without the constraints, the network struggles to learn the object location \& orientation even after 70k iters.

\textbf{Manipulating Latent Space for Design Recommendations from Database.} In the previous sub-section we showed our model's ability to generate indoor scenes given room layouts. However, apart from AI-synthesized recommendations one might also wish to obtain recommendations from human interior designers. This can be automated by having a large database curated using hand-designed interior rooms and then given an empty floor-plan by the user, recommend ``appropriate" designs from this database. However, the current practice is for the user to manually browse through configurations to find a match \cite{perez2019amzshowroom,havenly}. Instead, through our learnt latent space, we can use our model to suggest the best designs from the database to choose from. To simulate this, we consider the 3D-FRONT training set as our database. We use our learnt GNN encoder to convert each of these scenes into their corresponding latent code and store in the database. Then, given an empty room layout $\tilde{G}_R$ we retrieve the latent code with the highest likelihood under our autoregressive prior \eqref{eq: autoregressive model} (conditioned on $\tilde{G}_R$ and its type).\footnote{We evaluate \eqref{eq: optimal ordering} to compute the ordering $\pi$ of the latents for evaluating its likelihood under the autoregressive prior. More details in Appendix.} Finally, we pass this latent through the GNN decoder along with $\tilde{G}_R$. This ensures that the empty room $\tilde{G}_R$ is furnished according to the design of the retrieved scene while respecting the constraints imposed by $\tilde{G}_R$ e.g. furniture items should not go outside the walls. Figure \ref{fig: database retrieval} shows results for this experiment.

\begin{figure}[t]
  \centering
  \includegraphics[width=\linewidth]{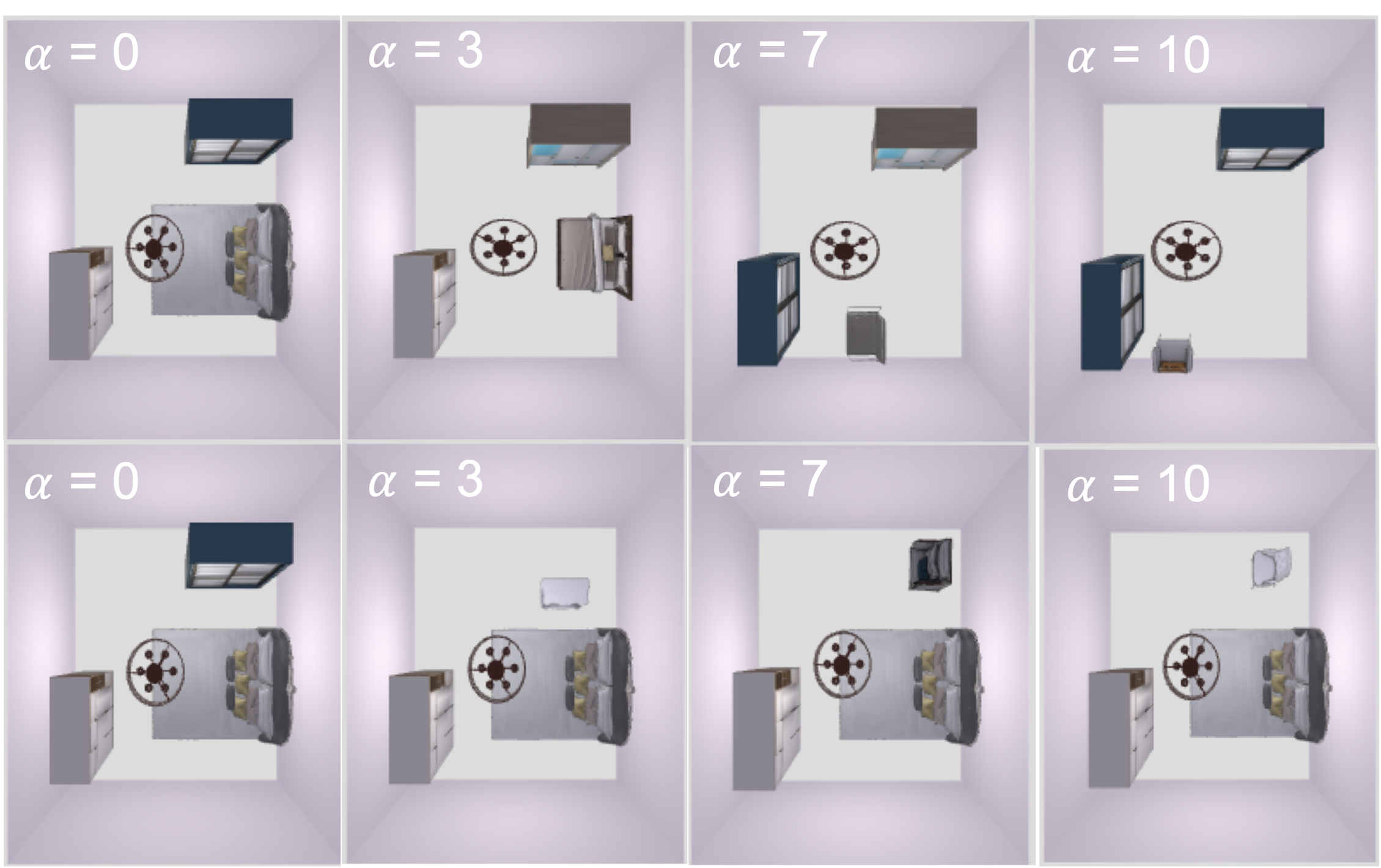}
   \caption{\textit{\textbf{Scene Editing.} Col 1 is a scene generated by our model. In row 1, we morph the bed into a chair by changing the $\alpha$ parameter as explained in text (\S\ref{sec. experiments}). In row 2, we morph the top cabinet into a chair.}}
   \label{Figure: Latent space traversals}
    \vspace{-3mm}
\end{figure}

\textbf{Scene Editing.} Our model also allows the user to traverse along the latent space to edit the scene content post-generation. Given a synthesized scene, we can convert a given furniture with label $c_1$ to another furniture labelled $c_2$ ensuring all other objects are approximately in the same geometric configuration. This is a non-trivial problem since depending on the room layout we also need to decide the size and orientation of the new furniture labelled $c_2$. In our formulation, this can be done by finding a latent direction $v:=\frac{\mu_2 - \mu_1}{||\mu_2 - \mu_1||_2}$ that transforms furniture from label $c_1 \rightarrow c_2$, where $\{\mu_{1}, \mu_2\}$ are the sample mean of the latent representations of furnitures with label $\{c_1, c_2\}$. Specifically, recall the GNN encoder in our model maps the input graph to a latent space with every furniture node having a separate latent code. We first compute the latent embeddings for all scenes in the training set. We then compute $\mu_{i}$ by averaging over all latent codes corresponding to furnitures with label $c_i$ across all scenes in the training set, where $i \in \{1, 2\}$. Now at inference time, we synthesize a scene $S$ with a furniture labelled $c_1$ using our autoregressive prior. Post-synthesis we can select the latent $\hat{z}$ corresponding to furniture $c_1$ and translate it to $\hat{z}' = \hat{z} + \alpha v$, where $\alpha$ controls the magnitude of morphing $c_1$ into $c_2$. This updated latent $\hat{z}'$ along with the latent codes of all other furniture items in $S$ are passed through the GNN decoder again. The end result, furniture $c_1$ gets morphed into $c_2$ while keeping the relative spatial arrangement of all other furniture's the same. This process is elucidated in Figure \ref{Figure: Latent space traversals}.

\section{Conclusion}
We have presented a latent-variable model for generating 3D indoor scenes given the room type and layout which is competitive with purely autoregressive models. Moreover, we show how a learnt latent space can be utilized to recommend designs from a database. In future work, we wish to explore more expressive non-linear autoregressive priors to improve generations; however the KL divergence cannot be computed analytically and the subsequent matching will no longer be quadratic. Finally, the matching procedure introduced in this paper can be potentially useful in other permutation-invariant domains like sets.

{\small
\bibliographystyle{dinat}
\bibliography{main}
}

\newpage\phantom{blabla}

\section{Appendix}

\input{appendix.tex}

\end{document}

%% file: math_commands.tex
\usepackage{amsmath,amsfonts,bm}
\usepackage{amssymb}

\def\1{\bm{1}}

\newcommand{\test}{\mathcal{D_{\mathrm{test}}}}

\DeclareMathAlphabet{\mathsfit}{\encodingdefault}{\sfdefault}{m}{sl}
\SetMathAlphabet{\mathsfit}{bold}{\encodingdefault}{\sfdefault}{bx}{n}

\def\gG{{\mathcal{G}}}

\def\gL{{\mathcal{L}}}

\def\gN{{\mathcal{N}}}

\def\sR{{\mathbb{R}}}

\DeclareMathOperator*{\argmin}{arg\,min}

%% file: appendix.tex
\subsection{Feature engineering}
As explained in \S\ref{subsection: indoor scene representation as a graph}, our indoor scene representation is an attributed graph. We now describe these attributes in detail. 

\begin{itemize}
	\item \textbf{Room node features $\boldsymbol {X_R}$},
	\begin{enumerate}
		\item \textbf{\textit{Node Type:}} A 4D 1-hot embedding of the node type. The four categories are wall, door, floor, window. 
		\item \textbf{\textit{Room Type:}} A 4D 1-hot embedding of the room type. The four categories are living room, bedroom, dining room, library. 
		\item \textbf{\textit{Location:}} A 6D vector representing the minimum and maximum vertex positions of the 3D bounding box corresponding to the room node.
		\item \textbf{\textit{Normal:}} A 3D vector representing the direction of the normal vector corresponding to the room node.
	\end{enumerate}

	\item \textbf{Furniture node features $\boldsymbol {X_F}$},
	\begin{enumerate}
		\item \textbf{\textit{Super category:}} A 7D 1-hot embedding of the node type. The  seven categories are Cabinet/Shelf, Bed, Chair, Table, Sofa, Pier/Stool, Lighting. 
		\item \textbf{\textit{Shape:}} A 1024D embedding of the 3D shape descriptor obtained by processing a 3D point cloud of the furniture item through PointNet.
		\item \textbf{\textit{Location:}} A 3D vector representing the centroid of the furniture item.
		\item \textbf{\textit{Orientation:}} A 3D vector representing the direction the ``front" side of the furniture is facing.
		\item \textbf{\textit{Size:}} A 3D vector representing the dimensions of the furniture along each axis.
	\end{enumerate}

	\item \textbf{Furniture-furniture edge features $\boldsymbol {X_{FF}}$}, let us arbitrarily chose a furniture-furniture edge and label its source node as $s$ with receiver node as $r$. We will describe the features for this edge.
	\begin{enumerate}
		\item \textbf{\textit{Center-center distance:}} A scalar representing the distance between the centroids of furniture $s$ and $r$. 
		\item \textbf{\textit{Relative orientation:}} A scalar representing the signed dot product between the orientation feature of $s$ and $r$.
		\item \textbf{\textit{Center-center orientation:}} A 3D vector representing the unit vector connecting the centroids of $r$ and $s$. 
		\item \textbf{\textit{Orientation:}} A 3D vector representing the direction the ``front" side of the furniture is facing.
		\item \textbf{\textit{Bbox-bbox distance:}} The shortest distance between the corners of the bounding box of $s$ and $r$.
	\end{enumerate}

	\item \textbf{Room-furniture edge features $\boldsymbol {X_{RF}}$}, let us arbitrarily chose a room-furniture edge and label its source node as $s$ with receiver node as $r$. We will describe the features for this edge.
	\begin{enumerate}
		\item \textbf{\textit{Center-room distance:}} A scalar representing the distance between the centroid of furniture $r$ and room node $s$. This is computed in 2D as a point to line distance. Every wall/window/door can be treated as a line segment in 2D. 
		\item \textbf{\textit{Center-room center:}} A scalar representing the  distance between the centroids of $r$ and $s$.  
		\item \textbf{\textit{Bbox-room dist}} A scalar representing the shortest distance between corners of the bounding box of furniture item $r$ to the room node $s$. This is computed in 2D as a point-to-line distance.
		\item \textbf{\textit{Bbox-room center}} A scalar representing the shortest distance between corners of the bounding box of furniture item $r$ to the centroid of the room node $s$.
		\item \textbf{\textit{Relative orientation:}} A signed dot product between the normal of $s$ with the orientation attribute of $r$. 	
	\end{enumerate}
	
	\item \textbf{Room-room edge features $\boldsymbol {X_{RR}}$}, let us arbitrarily chose a room-room edge and label its source node as $s$ with receiver node as $r$. We will describe the features for this edge.
	\begin{enumerate}
		\item \textbf{\textit{Center-center distance:}} A scalar representing the distance between the centroids of room nodes $s$ and $r$. 
		\item \textbf{\textit{Relative orientation:}} A scalar representing the signed dot product between the normal vectors of $s$ and $r$.
		\item \textbf{\textit{Longest distance:}} A scalar representing the longest distance between the corners of the bounding boxes of $r$ and $s$.
		\item \textbf{\textit{Shortest distance:}} A scalar representing the shortest distance between the corners of the bounding boxes of $r$ and $s$.
	\end{enumerate}
\end{itemize}

\subsection{Architecture Details}

\subsubsection{Message-Passing Graph Neural Network (MP-GNN)}
\label{subsubsection: Message-Passing Graph Neural Network}
Here we describe the message-passing mechanism employed in the Graph Neural Networks used in this work. The design of the MP-GNN is inspired by the work of \cite{mavroudi2020representation}, where each layer $l=1 \ldots,L$ of the MP-GNN maps a graph $G^{l-1}$ to another graph $G^l$ by updating the graph’s node and edge features. Specifically, let $h_i^l$ and $h_{ij}^l$  denote the features of node $i$ and edge $(i,j)$ of graph $G^l$, respectively. Let the input to the network be the graph $G^0=G$, so that $h_i^0$ and $h_{ij}^0$ denote the node features and edge features, respectively. At each iteration of node and edge refinement, the MP-GNN: (1) adapts the scalar edge weights by using an attention mechanism; (2) updates the edge attributes depending on the edge type, the attention-based edge weights, the attributes of the connected nodes and the previous edge attribute; and (3) updates the node attribute by aggregating attributes from incoming edges , as described next. 

\textit{Computing attention coefficients:} At each step $l$ of refinement, we update the scalar edge weights by computing an attention coefficient $a_{ij}^l$ that measures the relevance of node $j$ for node $i$ as follows:
\begin{equation}
	\begin{aligned}
				\gamma_{ij}^l = \rho(w_a^{\epsilon_{ij}\top} [W_r^{\nu_i} h_i^l ; W_s^{\nu_j} h_j^l ; W_{rs}^{\epsilon_{ij}} h_{ij}^l] \\ a_{ij}^l=\frac{\exp(\gamma_{ij}^l)}{\sum_{k\in N_{\epsilon}(i)}\exp(\gamma_{ik}^l)},   \qquad \forall (i,j) \in E.
		\end{aligned}
\end{equation}
In the first equation, each edge $(i,j)$ of $G_l =(V,E,X_l)$ is associated with a score, $\gamma^l_{ij}$, obtained by applying a nonlinearity $\rho$ (e.g., a leaky ReLU) to the dot product between a vector of attention weights $w_a^{\epsilon_{ij}}$ for edge type $\epsilon_{ij} \in \{RR,RF,FF\}$, and the concatenation of five vectors: (1) the receiver node feature $h_i^l \in \mathbb{R}^{d_i}$ weighted by a matrix $W_r^{\nu_i}$ for node type $\nu_i \in \{R, F\}$, (2) the sender node feature $h_j^l$ weighted by a matrix $W_s^{v_j}$ for node type $v_j$, and (3) the receiver-sender edge feature $h_{ij}^l$ weighted by a matrix $W_{rs}^{\epsilon_{ij}}$ for edge type $\epsilon_{ij}$. All weight matrices are also indexed by the GNN layer $l$ which is suppressed to prevent notational clutter.

\textit{Updating the node and edge features:} At each step $l$ of refinement, we update the edge and node features as: 
\begin{equation}
	\begin{aligned}
			h_{ij}^{l} &=  a_{ij}^l (U_{edge}^{\epsilon_{ij}}W_{s}^{\nu_j} h_j^{l-1}+W_{rs}^{\epsilon_{ij}}  h_{ij}^{l-1}),  &  \forall (i,j)\in E \\
			h_i^l &= \rho(h_i^{l-1}+ \sum_{k\in N_{\epsilon_{ij}}(i)} 
			U_{node}^{\epsilon_{ij}} W_sh_{ik}^{l}), & \forall i \in V.
		\end{aligned}
\end{equation}
The first equation updates the features of edge $(i,j)$ by taking a weighted combination of the features of node $j$ and the features of edge $(i,j)$ in the previous layer, with weight matrices $W_s^{\nu_j}$ and $W_{rs}^{\epsilon_{ij}}$, respectively. The matrix $U_{edge}^{\epsilon_{ij}}$ is a learnable projection matrix to change the dimensionality of the first term (transformed node feature) with that of the second term (transformed edge feature). 

The second equation updates the feature of node $i$ as the sum of the feature of node $i$ from the previous layer (a residual connection) and the sum of the features of all edges connected to node $i$ after applying a non-linearity $\rho$, such as a ReLU or leaky-ReLU. Here, $U_{node}^{\epsilon_{ij}}$ denotes a learnable projection matrix for edge type $\epsilon_{ij}$.

All weight matrices are also indexed by the GNN layer $l$ which is suppressed to prevent notational clutter. 

\subsubsection{Encoder, Decoder and Room Aggregator networks}
We employ three MP-GNNs in this work. One for the VAE encoder, one for the decoder and one for the Room Aggregator component of the prior network. Each of these MP-GNNs are three layered graph neural networks. In Table \ref{Table: encoder architecture}, \ref{Table: decoder architecture}, \ref{Table: room_agg architecture} we summarize the architecture of the Encoder, Decoder and Room Aggregator networks respectively. 

\begin{table}
\caption{\textit{Encoder architecture. The network parameters are as defined in Section \ref{subsubsection: Message-Passing Graph Neural Network}}}
	\centering
	\includegraphics[width=0.9\linewidth]{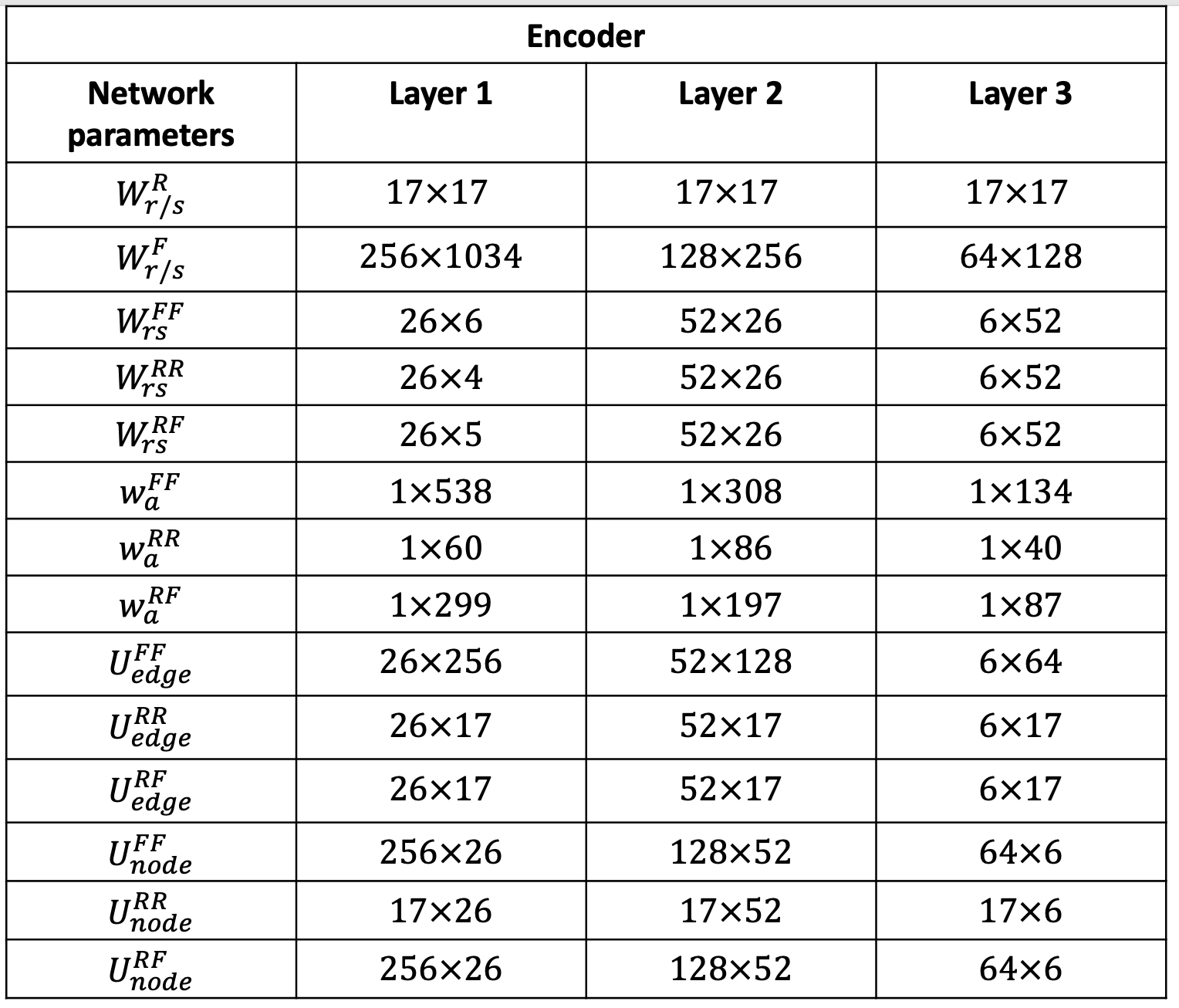}
	\label{Table: encoder architecture}
\end{table}

\begin{table}
	\centering
	\caption{\textit{Decoder architecture. The network parameters are as defined in Section \ref{subsubsection: Message-Passing Graph Neural Network}}}
	\includegraphics[width=0.9\linewidth]{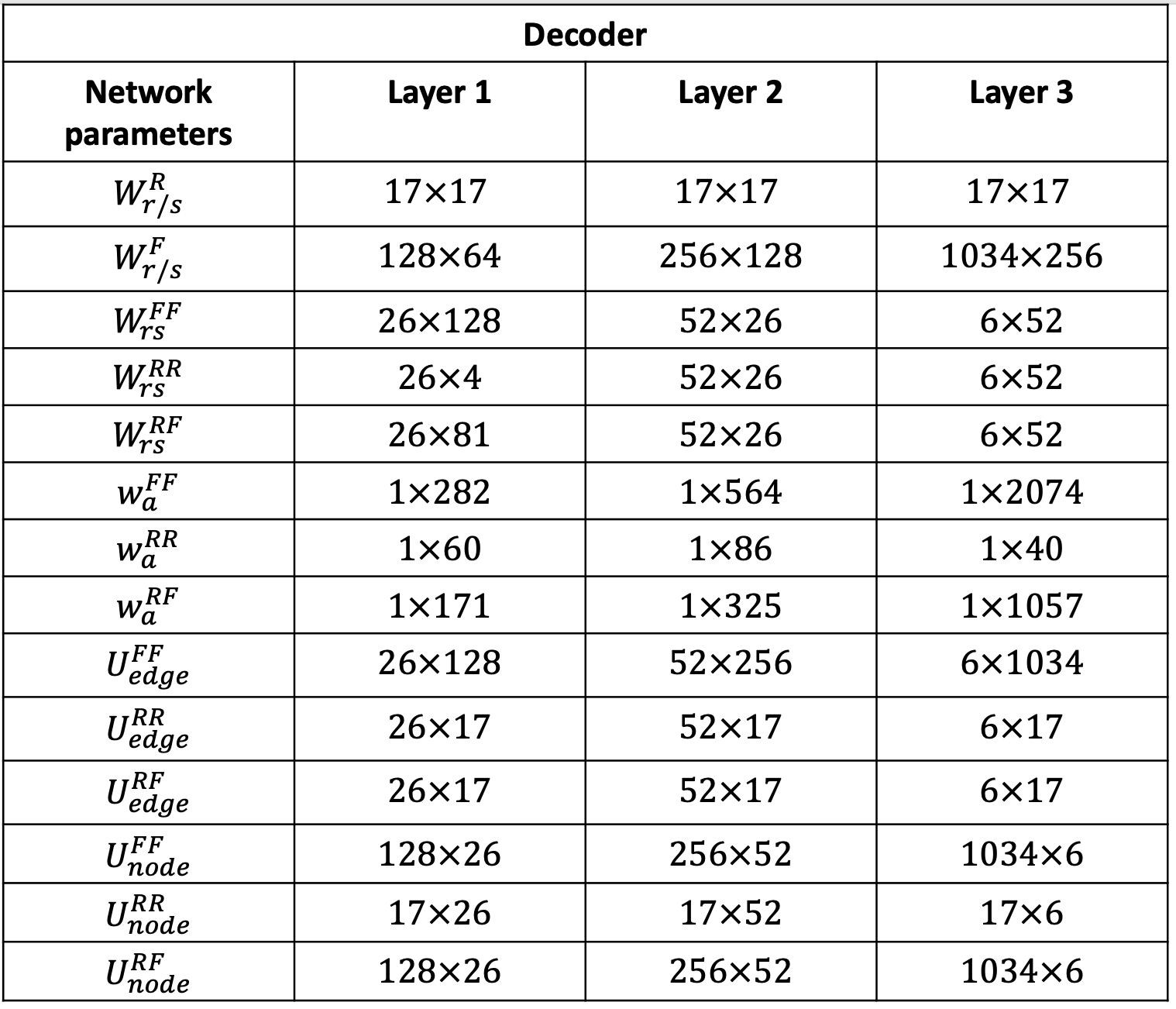}
	\label{Table: decoder architecture}
\end{table}

\begin{table}
	\caption{\textit{Room Aggregator architecture. This network only processes the room subgraph $G_R$ and hence all the weight matrices associated with furniture nodes are absent. The network parameters are as defined in \S \ref{subsubsection: Message-Passing Graph Neural Network}}}
	\centering
	\includegraphics[width=0.9\linewidth]{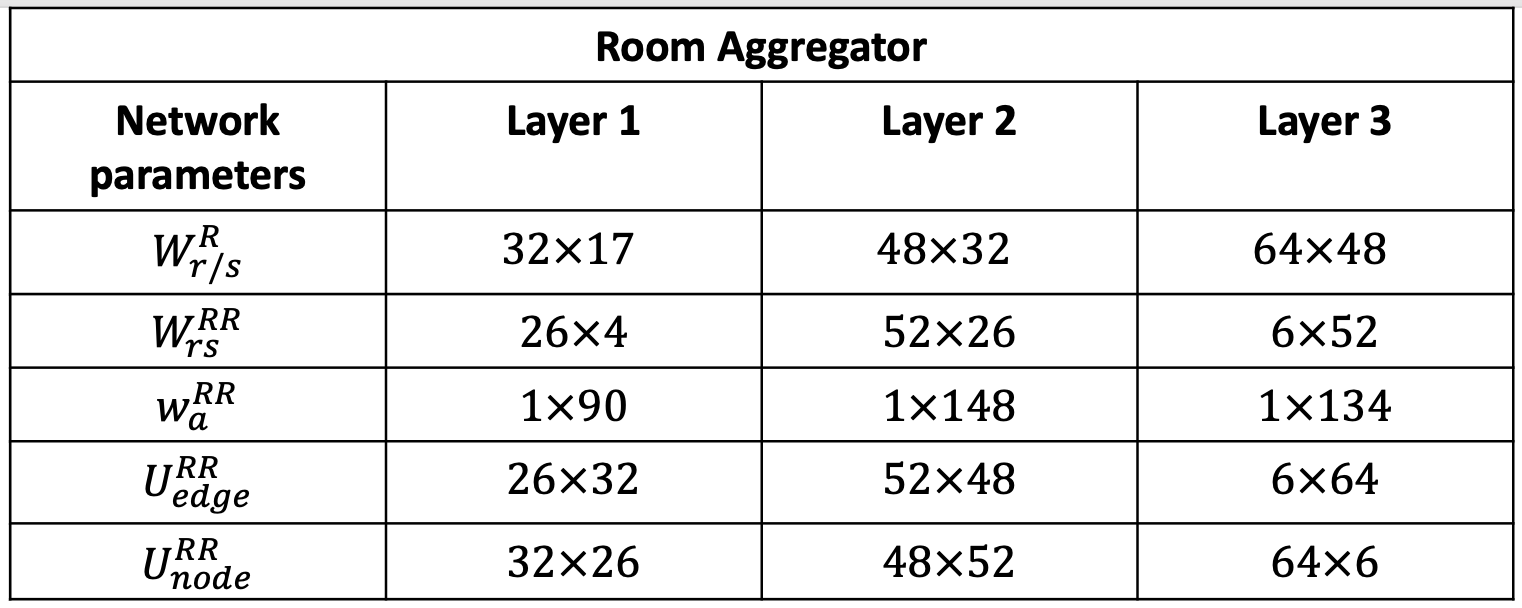}
	\label{Table: room_agg architecture}
\end{table}

The RNN Prior network (second component of our Graph Prior)  is implemented by a single layer Gated Recurrent Unit (GRU) with hidden vector of size $64 \times 64$ (the latent space is $64$D).

\subsection{The Reconstruction Loss of the ELBO}
The reconstruction loss is the first term in the ELBO \eqref{eq: ELBO}. 
\[\mathbb{E}_{Z \sim q_{\phi}(Z \mid G, T, n_F)} [\log p_{\theta'}(G_F \mid  Z,  n_F, G_R, T)]\]

In subsection \ref{subsection: proposed generative model} we described the distribution $p_{\theta'}(G_F \mid  Z,  n_F, G_R, T)$ as 
\begin{equation*}
	\begin{aligned}
		&p_{\theta'} (G_F \mid Z, n_F, G_R, T) \\=  &\prod_{i=1}^{n_F}p_{\theta'}(shape_i \mid Z, G_R, T) p_{\theta'}(orien_i \mid Z, G_R, T) \\
		& p_{\theta'}(loc_i \mid Z, G_R, T)p_{\theta'}(size_i \mid shape_i) p_{\theta'}(cat_i \mid shape_i).
	\end{aligned}
\end{equation*}
Here $shape_i, orien_i, loc_i, size_i$ and $cat_i$ refer to the ground truth values of the 3D shape descriptor (PointNet features), orientation, location, size and category  respectively for furniture node $i$ in $G_F$. 

We will now describe each of these distributions in detail along with the corresponding loss term. For the normal and lognormal distributions used we assume the variance parameter is a fixed constant equal to  $1$ and do not learn them. 

\begin{equation*}
	\begin{aligned}
		p_{\theta'}(shape_i \mid Z, G_R, T) &= \gN(\mu^{shape}_{\theta'}(Z, G_R, T), const.) \\
		\log p_{\theta'}(shape_i \mid Z, G_R, T) &= -\left(\mu^{shape}_{\theta'}(Z, G_R, T) - shape_i \right)^2
	\end{aligned}
\end{equation*}

\begin{equation*}
	\begin{aligned}
		p_{\theta'}(orient_i \mid Z, G_R, T) &= Categorical(\Theta^{orient}_{\theta'}(Z, G_R, T)) \\
		\log p_{\theta'}(orient_i \mid Z, G_R, T) &= -H(orient_i, \Theta^{orient}_{\theta'}(Z, G_R, T))
	\end{aligned}
\end{equation*}
$orient_i$ is a $4$D $1$-hot embedding of the ground truth orientation and $\Theta$ is the parameter of the categorical distribution. $H(.,.)$ denotes the cross entropy between two distributions. 

\begin{equation*}
	\begin{aligned}
		p_{\theta'}(loc_i \mid Z, G_R, T) &= \gN(\mu^{loc}_{\theta'}(Z, G_R, T), const.) \\
		\log p_{\theta'}(loc_i \mid Z, G_R, T) &= -\left(\mu^{loc}_{\theta'}(Z, G_R, T) - loc_i \right)^2
	\end{aligned}
\end{equation*}

\begin{equation*}
	\begin{aligned}
		p_{\theta'}(size_i \mid shape_i) &= LogNormal(\mu^{size}_{\theta'}(\mu^{shape}_{\theta'}(Z, G_R, T)), const.) \\
		\log p_{\theta'}(size_i \mid shape_i) &= -\left(\mu^{size}_{\theta'}(\mu^{shape}_{\theta'}(Z, G_R, T)) - \ln size_i \right)^2
	\end{aligned}
\end{equation*}

\begin{equation*}
	\begin{aligned}
		p_{\theta'}(cat_i \mid shape_i) &= Categorical(\Theta^{cat}_{\theta'}(\mu^{shape}_{\theta'}(Z, G_R, T))) \\
		\log p_{\theta'}(cat_i \mid shape_i) &= -H(cat_i, \Theta^{cat}_{\theta'}(\mu^{shape}_{\theta'}(Z, G_R, T)))
	\end{aligned}
\end{equation*}

Each furniture node feature in the output of the MP-GNN Decoder is individually processed using a multi-layer Perceptron (MLP). Specifically, let $\hat{h}_i^L$ be the output feature of the decoder for furniture node $i$. In our experiments $\hat{h}_i^L \in \mathbb{R}^{1034}$. 

\[\mu^{shape}_{\theta'}(Z, G_R, T) = \textrm{Linear}_{\theta'}(1034, 1024)\]
\[\Theta^{orient}_{\theta'}(Z, G_R, T)) = \textrm{SoftMax}(\textrm{Linear}_{\theta'}(1034, 4))\]

\[\mu^{loc}_{\theta'}(Z, G_R, T) = \textrm{Linear}_{\theta'}(1034, 3)\]
\[\mu^{size}_{\theta'} = \textrm{Linear}_{\theta'}(1024, 512) \rightarrow \textrm{ ReLU } \rightarrow \textrm{Linear}_{\theta'}(512, 3)\]
This is a three-layer MLP with ReLU activations and $1024$ input neurons since the input is the mean of the shape features $\mu^{shape}_{\theta'}(Z, G_R, T))$. 

\[\Theta^{cat}_{\theta'}= \textrm{Linear}_{\theta'}(1024, 512) \rightarrow \textrm{ ReLU } \rightarrow \textrm{Linear}_{\theta'}(512, 7)\]

The output is $7$D since there are $7$ super-categories in our dataset, Cabinet/Shelf, Bed, Chair, Table, Sofa, Pier/Stool, Lighting. 

$\textrm{Linear}_{\theta'}(x, y)$ denotes a linear layer with $x$ input neurons and $y$ input neurons. Recall $\theta'$ denotes all the parameters of the Decoder including the GNN and the output MLPs. 

\subsection{Computing the KL divergence term}

\subsubsection{Derivation for \eqref{eq. KL div. analytical form}}
\label{subsection: Derivation for KL term}
Rewriting \eqref{eq: optimal ordering}
\begin{equation}
	\pi^* = \argmin_\pi KL( q_{\phi}(Z \mid G, T, n_F) \mid\mid p_{\theta''}(Z \mid G_R, \pi, n_F, T))
	\label{eq: optimal ordering appendix}
\end{equation}

For simplicity let us denote the two distributions as
\begin{equation*}
	\begin{aligned}
		q_{\phi}(Z \mid G, T, n_F) &= \gN(\mu_0, \Sigma_0) \\ p_{\theta''}(Z \mid G_R, \pi, n_F, T) &= \gN(\tilde{\pi}\mu_1,\tilde{\pi}\Sigma_1\tilde{\pi}^T),
	\end{aligned}	
\end{equation*}
where $\tilde{\pi} = \pi \otimes I_{d_F \times d_F}$. \\
The KL divergence between two Gaussian distributions is known analytically,
\begin{equation}
	\begin{aligned}
		& KL( q_{\phi}(Z \mid G, T, n_F) \mid\mid p_{\theta''}(Z \mid G_R, \pi, n_F, T)) \\
		&= \frac{1}{2}\left( Tr(\tilde{\pi}\Sigma_1^{-1}\tilde{\pi}^T \Sigma_0) + (\tilde{\pi}\mu_1 - \mu_0)^T\tilde{\pi}\Sigma_1^{-1}\tilde{\pi}^T(\tilde{\pi}\mu_1 - \mu_0) \right) \\
		&\quad  + \frac{1}{2}\left( \ln\left[\frac{det(\tilde{\pi}\Sigma_1\tilde{\pi}^T)}{det(\Sigma_0)}\right] - n_Fd_F \right) \\
		&\equiv \left( Tr(\tilde{\pi}\Sigma_1^{-1}\tilde{\pi}^T \Sigma_0) + (\tilde{\pi}\mu_1 - \mu_0)^T\tilde{\pi}\Sigma_1^{-1}\tilde{\pi}^T(\tilde{\pi}\mu_1 - \mu_0) \right) \\
		&= Tr(\Sigma_1^{-1}\tilde{\pi}^T \Sigma_0\tilde{\pi}) + \mu_1^T 
		\Sigma^{-1}\mu_1 - \mu_1^T \Sigma_1^{-1} \tilde{\pi}^T\mu_0 \\& \quad- \mu_0^T\tilde{\pi}\Sigma_1^{-1}\mu_1 + \mu_0^T\tilde{\pi}\Sigma_1^{-1}\tilde{\pi}^T\mu_0\\
		&\equiv Tr(\Sigma_1^{-1}\tilde{\pi}^T \Sigma_0\tilde{\pi}) - \mu_1^T \Sigma_1^{-1} \tilde{\pi}^T\mu_0 - \mu_0^T\tilde{\pi}\Sigma_1^{-1}\mu_1 \\& \quad+ \mu_0^T\tilde{\pi}\Sigma_1^{-1}\tilde{\pi}^T\mu_0\\ 
		&\equiv Tr(\Sigma_1^{-1}\tilde{\pi}^T \Sigma_0\tilde{\pi}) - 2\mu_1^T \Sigma_1^{-1} \tilde{\pi}^T\mu_0 + \mu_0^T\tilde{\pi}\Sigma_1^{-1}\tilde{\pi}^T\mu_0\\
	\end{aligned}
\label{eq: matching derivation}
\end{equation}
The third equivalence is obtained by observing that the constant $n_Fd_F$ (which is the dimension of the support of the two Gaussian distributions) does not affect the solution of \eqref{eq: optimal ordering appendix} and that permuting the rows and column of a matrix by the same permutation does not change its determinant and hence the minimizer $\pi^*$ will also not depend on this term. By similar reasoning, we also ignore the factor of $\frac{1}{2}$. In the fifth equivalence we again ignore terms that don't affect $\pi^*$. Using the cyclic property of the $Trace$ operator we can rewrite the last term on the RHS of \eqref{eq: matching derivation} as 
\[\mu_0^T\tilde{\pi}\Sigma_1^{-1}\tilde{\pi}^T\mu_0 = Tr\left(\mu_0^T\tilde{\pi}\Sigma_1^{-1}\tilde{\pi}^T\mu_0\right) = Tr\left(\Sigma_1^{-1}\tilde{\pi}^T\mu_0\mu_0^T\tilde{\pi}\right)\]
Substituting this results in \eqref{eq: matching derivation} we obtain the desired result. 
\begin{equation*}
	\begin{aligned}
		& KL( q_{\phi}(Z \mid G, T, n_F) \mid\mid p_{\theta''}(Z \mid G_R, \pi, n_F, T)) \\
		&\equiv Tr(\Sigma_1^{-1}\tilde{\pi}^T \Sigma_0\tilde{\pi}) - 2\mu_1^T \Sigma_1^{-1} \tilde{\pi}^T\mu_0 + Tr\left(\Sigma_1^{-1}\tilde{\pi}^T\mu_0\mu_0^T\tilde{\pi}\right)\\
		&= Tr(\Sigma_1^{-1}\tilde{\pi}^T \left[ \Sigma_0 + \mu_0\mu_0^T \right] \tilde{\pi}) - 2\mu_1^T \Sigma_1^{-1} \tilde{\pi}^T\mu_0 \\
		&= Tr(\Sigma_1^{-1}\tilde{\pi}^T \left[ \Sigma_0 + \mu_0\mu_0^T \right] \tilde{\pi}) - 2 Tr\left(\mu_1^T \Sigma_1^{-1} \tilde{\pi}^T\mu_0\right) \\
		&= Tr(\Sigma_1^{-1}\tilde{\pi}^T \left[ \Sigma_0 + \mu_0\mu_0^T \right] \tilde{\pi}) - 2 Tr\left(\tilde{\pi}^T\mu_0\mu_1^T \Sigma_1^{-1} \right) \\
	\end{aligned}
\end{equation*}
We again used the cyclic property of the $Trace$ operator for the last equality. This concludes our derivation for \eqref{eq. KL div. analytical form}.

\subsubsection{Solving \eqref{eq. KL div. analytical form} using the FAQ approximation}.
\label{subsection: Solving using the FAQ approximation}
 The FAQ algorithm proceeds as follows, 
\begin{enumerate}
	\item \textbf{Choose an initial point:} We choose the uninformative $\pi_0 = \frac{1.1^T}{n_F}$ as the initial estimate for the algorithm. Note, here $\pi^0$ is a doubly stochastic matrix and not a permutation matrix.
	\item \textbf{Find a local solution:} In each iteration $i$, we linearize the objective at the current iterate $\pi^i$
	\[\tilde{f^i}(\pi) := f(\pi^i) + Tr\left[\nabla f(\pi^i)^T(\pi - \pi^i)\right] \]
	Here, $f(\pi) = Tr\left(\Sigma_1^{-1} \tilde{\pi} \left[ \Sigma_0 + \mu_0\mu_0^T \right]\tilde{\pi}^T\right)  - 2Tr\left(\tilde{\pi} \mu_0\mu_1^T\Sigma_1^{-1}\right)$. 
	We then solve the following subproblem
	\begin{equation}
			\begin{aligned}
					\min_{\pi}  &Tr(\nabla f(\pi^i)^T(\pi) \\
					\textrm{s.t. } & \pi \in D
				\end{aligned}
		\label{eq: FW subproblem}
		\end{equation}
   Here $D$ is the set of all doubly stochastic matrices. Notice \eqref{eq: FW subproblem} is just a Linear Assignment Problem and can be efficiently solved by the Hungarian Algorithm. Let $q^i$ denote the $\argmin$ of \eqref{eq: FW subproblem}.
   \item \textbf{Choose the next iterate:} Finally choose $\pi^{i+1} = \alpha \pi^i + (1 - \alpha) q^i$, where $\alpha \in [0,1]$. Here $\alpha$ is chosen by solve the 1D optimization problem analytically. 
   \begin{equation}
	   	\min_{\alpha} f(\alpha \pi^i + (1 - \alpha) q^i) \quad \textrm{s.t. } \alpha \in [0, 1]
	   	\end{equation}
   	
	\item \textbf{Repeat steps $2$ and $3$ untill convergence.} Steps 2-3 are repeated iteratively untill some termination criteria is met, say $||\pi^i - \pi^{i-1} ||_F< \epsilon$. In practice, we do not run steps 2-3 untill convergence but terminate after just $1$ iteration of the FAQ algorithm. Running for longer iterations did not result in any significant gains in performance.

   \item \textbf{Project onto the set of permutation matrices.} Upon termination, the final solution is obtained by projecting $\pi^{final}$ onto the space of permutation matrices by solving $\min_{\pi \in P} -Tr(\pi^{final}\pi^T)$. Here $P$ is the set of all permutation matrices of size $n_F$. This is against solved by the Hungarian Algorithm. 
\end{enumerate}

In step 2, we need to compute the gradient of $f(\pi^i)$. This can be done analytically. We will start our derivation by stating some facts. First, the gradient of the dot product of two matrices $A$ and $X$ with respect to $X$ is,
\begin{equation}
	\frac{d}{dX}\langle A, X \rangle = A
	\label{eq: derivative of dot product}
\end{equation}

Second, for every linear operator $M$, its adjoint operator $M^{\dagger}$ is defined such that 
\begin{equation}
	\langle A, M(X) \rangle = \langle M^{\dagger}(A), X \rangle
	\label{eq: definittin of adjojint}
\end{equation}

Recall,
\[f(\pi) = Tr\left(\Sigma_1^{-1} \tilde{\pi} \left[ \Sigma_0 + \mu_0\mu_0^T \right]\tilde{\pi}^T\right)  - 2Tr\left(\tilde{\pi} \mu_0\mu_1^T\Sigma_1^{-1}\right)\]. 

We will fist look at the second term, 
\begin{equation}
	\begin{aligned}
		Tr\left(\tilde{\pi} \mu_0\mu_1^T\Sigma_1^{-1}\right) &= \langle \Sigma_1^{-1} \mu_1 \mu_0^T, M(\pi) \rangle \\
		&= \sum_{ij}^{n_F \times n_F} \pi_{ij} Tr\left(\left[\Sigma_1^{-1} \mu_1 \mu_0^T\right]_{ij} \right) \\
		&=  \langle M^{\dagger}\left(\Sigma_1^{-1} \mu_1 \mu_0^T\right), \pi \rangle \\		
	\end{aligned}
\label{eq. deriving the adjoint}
\end{equation}
Here, $M(\pi) := \pi \otimes I_{d_F \times d_F}$. Given a matrix $A \in \mathbb{R}^{n_Fd_F \times n_Fd_F}$, we define the operation $\left[A \right]_{ij}$ (used in the second equality) as follows. First divide the matrix $A$ into non-overlapping blocks of size $d_F \times d_F$, there are $n_F \times n_F$ such blocks.  Now $\left[A\right]_{ij}$ denotes the $ij^{th}$ block. In the last equality, we defined the adjoint $L^{\dagger}(A)$ as $\hat{A}$. Here $\hat{A}$ is a $n_F \times n_F$ matrix obtained from $A$ whose $ij^{th}$ entry is, 
\[\hat{A}_{ij} = Tr\left(\left[A \right]_{ij}\right).\].

Combining \eqref{eq. deriving the adjoint} with \eqref{eq: derivative of dot product} we conclude 
\begin{equation}
	\frac{d}{d\pi} Tr\left(\tilde{\pi} \mu_0\mu_1^T\Sigma_1^{-1}\right) = M^{\dagger}\left(\Sigma_1^{-1} \mu_1 \mu_0^T\right)
\end{equation}

The gradient of the first term is calculated similarly, 

\begin{equation}
	\begin{aligned}
		Tr\left(\Sigma_1^{-1} \tilde{\pi} \left[ \Sigma_0 + \mu_0\mu_0^T \right]\tilde{\pi}^T\right) &= \langle \Sigma_1^{-1} \tilde{\pi} \left[ \Sigma_0 + \mu_0\mu_0^T \right],M(\pi)  \rangle \\
		\frac{d}{d\pi}Tr\left(\Sigma_1^{-1} \tilde{\pi} \left[ \Sigma_0 + \mu_0\mu_0^T \right]\tilde{\pi}^T\right) &= 2 M^{\dagger}(\Sigma_1^{-1} \tilde{\pi} \left[ \Sigma_0 + \mu_0\mu_0^T \right])
	\end{aligned}
\end{equation}
Here $M(\pi)$ is again defined as $ \pi \otimes I_{d_F \times d_F}$. Putting it all together,

\begin{equation}
	\begin{aligned}
		\nabla f(\pi) = 2M^{\dagger}(\Sigma_1^{-1} \tilde{\pi} \left[ \Sigma_0 + \mu_0\mu_0^T \right]) - 2M^{\dagger}\left(\Sigma_1^{-1} \mu_1 \mu_0^T\right)
	\end{aligned}
\end{equation}

\subsection{Learning Under Constraints}
\label{appendix: learning under constraints}
Let furniture graph $G_F$ be the input to the encoder and $\tilde{G}_F$ be the furniture graph reconstructed by the decoder. Recall, $E_{FF}$ denotes the edges between furniture nodes in $G_F$ (\S \ref{subsection: indoor scene representation as a graph}) and $\tilde{G}_F$ has the same structure as $G_F$.  We employ the following constraints,

 \begin{itemize}
	 	\item \textbf{furniture-furniture distance constraint}: For every $(v_i, v_j ) \in E_{FF}$, we define there relative position as $c_{ij} = ||loc(v_i) - loc(v_j)||_2$, where $loc(v_i)$ denotes the $3$D centroid of furniture item $v_i$ from $G_F$. We define $c^{pred}_{ij} = ||loc^{pred}(v_i) - loc^{pred}(v_j)||_2$ as the relative distance computed from the corresponding location mean prediction by the decoder from $\tilde{G}_F$. Finally we define $d^{ij}_{FF} := MSE(c_{ij}, c^{pred}_{ij})$. Here $MSE$ refers to mean squared error.
	 	 
	 	\item \textbf{furniture-room distance constraint}: This constraint restricts the relative position of the predicted furniture items with the room nodes. 
	 	\begin{itemize}
		 		\item For the windows and doors, $d^{ij}_{RF}$ is defined analogously as \\ $d^{ij}_{RF} = MSE(c_{ij}, c^{pred}_{ij})$ where $(v_i, v_j) \in E_{RF}$, with the exception that for computing $c^{pred}_{ij}$ we use the ground truth room node location since they are not predicted by the decoder. 
		 		\item For walls, $c^{pred}_{ij}$ is computed as the signed distance between the $i^{th}$ wall node and the $j^{th}$ furniture centroid. This ensures the decoder predicts furniture items on the correct side of the wall. 
		 	\end{itemize}
	     \item \textbf{furniture-furniture relative orientation constraint}:
	 	 This constraint enforces the relative orientation between two predicted furniture centroid locations to be close to the ground truth. Specifically we define  $\forall (v_i, v_j) \in E_{FF}$ $o^{ij}_{FF} = orient(v_i, v_j)^Torient^{pred}(v_i, v_j)$. Here $orient(v_i, v_j)^T$ is the unit vector pointing in the direction $loc(v_i) - loc(v_j)$ computed from $G_F$. Similarly, $orient^{pred}(v_i, v_j)$ is the unit vector pointing in the direction $loc^{pred}(v_i) - loc^{pred}(v_j)$ computed from the reconstruction $\tilde{G}_F$. 
	 \end{itemize}

We now present the complete optimization objective as 
\begin{equation}
	\begin{aligned}
		&\max_{\theta', \theta'', \phi} \gL(\theta', \theta'', \phi) \\
		& s.t. \qquad \frac{1}{n} \sum_{i=1}^n  \left[\sum_{(v_i, v_j) \in E_{FF}} d^{ij}_{FF}\right] \leq \epsilon \\
		& s.t. \qquad \frac{1}{n} \sum_{i=1}^n  \left[\sum_{(v_i, v_j) \in E_{RF}} d^{ij}_{RF}\right] \leq \epsilon \\
		& s.t. \qquad \frac{1}{n} \sum_{i=1}^n  \left[\sum_{(v_i, v_j) \in E_{FF}} o^{ij}_{FF}\right] \geq 1 - \epsilon
		\end{aligned}
	\label{eq: constraints}
\end{equation}

Here $\epsilon$ is a user-defined hyperparameter that determines the strictness of enforcing these constraints\footnote{One could also have used a different $\epsilon_i$ per constraint.}.

 $\gL(\theta', \theta'', \phi) $ is as defined in \eqref{eq: overall ELBO} and $i$ is in iterator over the scene graphs in the training set. We employ the learning under constraints framework introduced by \cite{chamon2020probably} which results in a primal-dual saddle point optimization problem. For completeness, we will now describe this algorithm in detail. We begin by explicitly writing out the empirical lagrangian $\hat{\gL}_{\theta', \theta'', \phi, \lambda_1, \lambda_2, \lambda_3}$,

\begin{equation}
	\begin{aligned}
		g_1(\theta', \theta'', \phi) &:= \frac{1}{n} \sum_{i=1}^n  \left[\sum_{(v_i, v_j) \in E_{FF}} d^{ij}_{FF}\right] \\
		g_2(\theta', \theta'', \phi) &:= \frac{1}{n} \sum_{i=1}^n  \left[\sum_{(v_i, v_j) \in E_{RF}} d^{ij}_{RF}\right]\\
		g_3(\theta', \theta'', \phi) &:= \frac{1}{n} \sum_{i=1}^n  \left[\sum_{(v_i, v_j) \in E_{FF}} o^{ij}_{FF}\right] \\
		\hat{\gL}_{\theta', \theta'', \phi, \lambda_1, \lambda_2, \lambda_3} &:= \gL(\theta', \theta'', \phi) +  \lambda_3(1 - \epsilon - g_3(\theta', \theta'', \phi)) \\&+ \lambda_1(g_1(\theta', \theta'', \phi) - \epsilon) + \lambda_2(g_2(\theta', \theta'', \phi) - \epsilon)
	\end{aligned}
\label{eq: empirical lagrangian}
\end{equation}
Here $\lambda_1, \lambda_2, \lambda_3$ are the dual variables.   The empirical dual problem is then defined as, 
\[\hat{D}^* := \max_{\lambda_1, \lambda_2, \lambda_3} \min_{\theta', \theta'', \phi}\hat{\gL}_{\theta', \theta'', \phi, \lambda_1, \lambda_2, \lambda_3}.\]
It was shown in \cite{chamon2020probably} that a saddle-point optimization of this empirical dual will give an approximate solution to \eqref{eq: constraints}. Algorithm \ref{algorithm: constraints} describes the exact steps. At step $5$ we require a $\rho$-optimal minimizer to the empirical Lagrangian, in practice this is done by running the ADAM optimizer for one epoch. After each epoch $t$, the dual variables are updated depending on the slack evaluated with current parameters $(\theta'^{(t-1)}, \theta''^{(t-1)}, \phi^{(t-1)})$. At an intuitive level, the algorithm is similar to regularized optimization with adaptive Lagrange multipliers (the dual variables). In \eqref{eq: empirical lagrangian} each term after $\gL(\theta' , \theta'', \phi)$ can be thought of a regularizer (one corresponding to each constraint). The Langrange multipliers are updated in each epoch to enforce or relax the regularizer depending on whether the corresponding constraint is violated or satisfied. 

\begin{algorithm}
	\caption{Learning under constraints}
	\begin{algorithmic}[1]
		\Require \textit{Initializations:} $\theta'^{(0)}, \theta''^{(0)}, \phi^{(0)}$; \textit{Learning rate for dual variables} $\eta$; \textit{Number of steps} $T$ 
		\State $\lambda_1^{(0)} = 0$
		\State $\lambda_2^{(0)} = 0$
		\State$\lambda_3^{(0)} = 0$
		\For {$t = 1, \ldots, T$} \\
			\quad Obtain $\theta'^{(t-1)}, \theta''^{(t-1)}, \phi^{(t-1)}$ such that, \[ \hat{\gL}_{\theta', \theta'', \phi, \lambda_1, \lambda_2, \lambda_3} \leq \min_{\theta', \theta'', \phi} \hat{\gL}_{\theta', \theta'', \phi, \lambda_1, \lambda_2, \lambda_3} + \rho .\] \\
			
			\quad Update dual variables 
			\[\lambda_1 ^{(t)} = \left[ \lambda_1 ^{(t-1)} + \eta(g_1(\theta'^{(t- 1)}, \theta''^{(t- 1)}, \phi^{(t- 1)}) - \epsilon)  \right]_+\]
			\[\lambda_2 ^{(t)} = \left[ \lambda_2 ^{(t-1)} + \eta(g_2(\theta'^{(t- 1)}, \theta''^{(t- 1)}, \phi^{(t- 1)}) - \epsilon)  \right]_+\] 
			\[\lambda_3 ^{(t)} = \left[ \lambda_3 ^{(t-1)} + \eta(1 - \epsilon - g_3(\theta'^{(t- 1)}, \theta''^{(t- 1)}, \phi^{(t- 1)}))  \right]_+\]
		\EndFor
	\end{algorithmic}
\label{algorithm: constraints}
\end{algorithm}

\begin{figure*}[ht]
	\centering
	\includegraphics[width=1.0\linewidth]{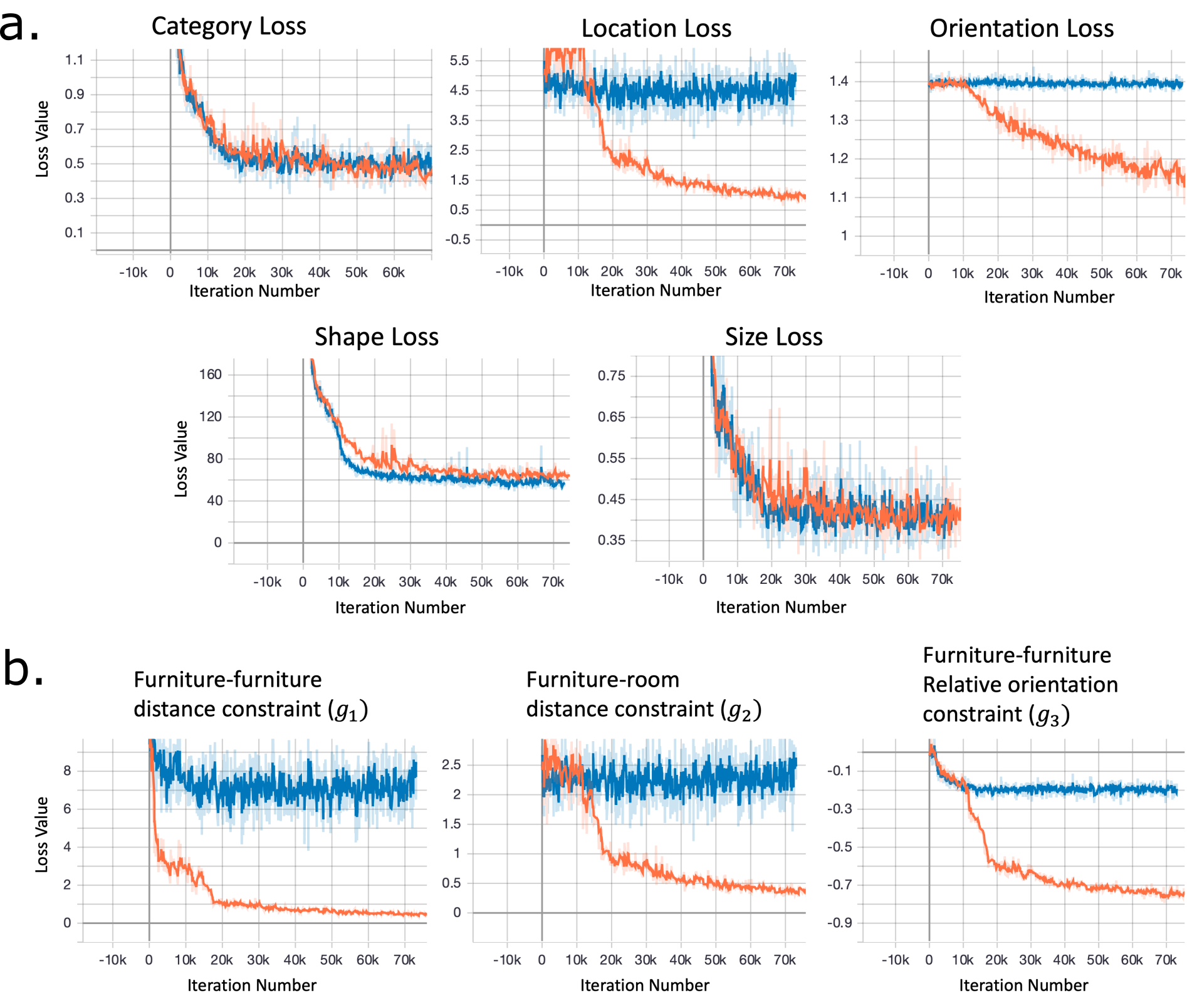}
	\setlength{\belowcaptionskip}{-10pt} 
	\caption{\textit{\textbf{Ablation studies showing the effect of constraints on training. The orange curve indicates learning under constraints whereas the blue curve indicates learning without any constraints}(a) Training curves for the terms in the reconstruction loss of the ELBO (first term in \eqref{eq: ELBO}; (b) Objective values of the constraints as training progresses. Note the x-axis is the number of iterations rather than epoch (one iteration is one batch processed).}}
	\label{fig: constraints}
\end{figure*}

In Figure \ref{fig: constraints}, we show training curves for ELBO objective \eqref{eq: ELBO} and the ELBO objective with constraints \eqref{eq: constraints} which clearly show the benefit of using constraints. Empirically on the 3D-FRONT dataset, getting good solutions is impossible (even after $900$ epochs ($70$k iterations)) without using constraints. The network performance without any constraints is comparable, to that with constraints, for the furniture category, shape and size loss terms. These are arguably much easier to learn than the orientation and position in $3$D. 

\newpage
\subsection{Diverse furniture layout recommendations for the same room layout}

\begin{figure*}[ht]
	\centering
	\includegraphics[width=1.0\linewidth]{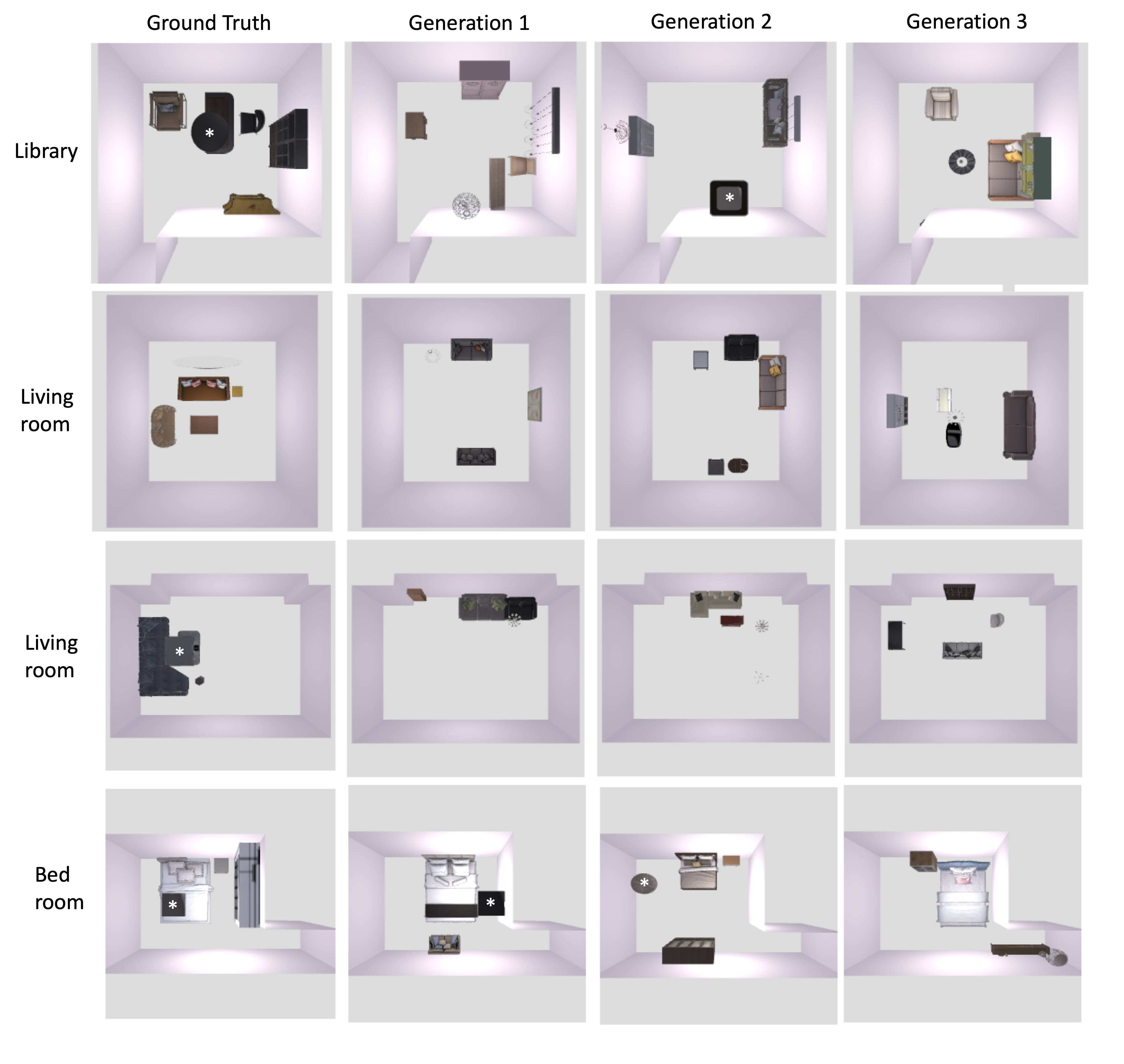}
	\setlength{\belowcaptionskip}{-10pt} 
	\caption{\textit{\textbf{Diverse furniture layout recommendations for the same room layout.} Each row depicts a specific floor-plan, row 1: library; row 2, 3: living room, row 4: bedroom. The first column is the ground truth design from the test set. The remaining columns are recommendations made by our proposed model. All rooms are top-down rendering of the scene. We mark ceiling lamps with a white asterisk in images where we believe it is hard to recognize from our top-down rendering. In row 1, column 4, the green furniture on top of the sofa is a overhead cabinet.}}	
	\label{fig: more examples for 2a}
\end{figure*}

More examples for Figure \ref{fig: proposed architecture}b in Figure \ref{fig: more examples for 2a}

\newpage
\subsection{Manipulating Latent Space for Design Recommendations
from Database.}

\begin{figure}[h]
	\centering
	\includegraphics[width=0.85\linewidth]{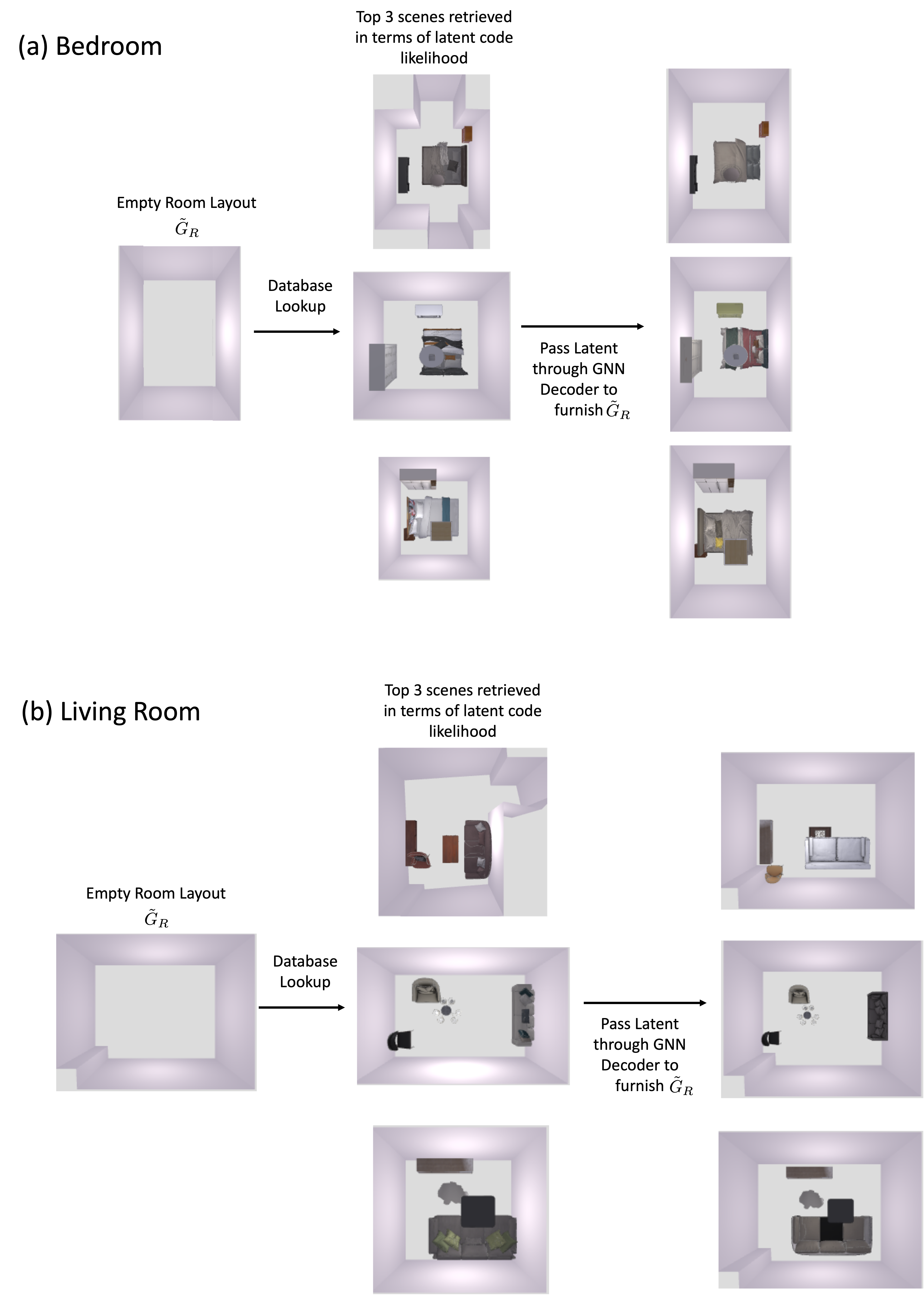}
	\caption{\textit{\textbf{Manipulating Latent Space for Design Recommendations
from Database.} We show results for two room types (a) bedroom and (b) living room. The GNN decoder ensures the synthesized design in $\tilde{G}_R$ has approximately the same spatial arrangement as the design of the retrieved scene. However in some cases (middle room in (a) and top room in (b), the bed and sofa are rotated respectively to ensure the satisfaction of constraints levied by the room layout, for example, sofa should be parallel to the closest wall.}}	
	\label{fig: more examples for database retrieval}
\end{figure}

In this subsection, we show more Examples for Figure \ref{fig: database retrieval} in Figure \ref{fig: more examples for database retrieval}. The procedure is as follows,

\begin{enumerate}
    \item Create a Database:
    \begin{itemize}
        \item Iterate over the training set.
        \item For each scene, pass it through the GNN encoder to obtain the latent code. Store both the scene and the corresponding latent code in the database.
    \end{itemize}
    \item Retrieving closest matched scenes
    \begin{itemize}
    \item Given an empty room layout $\tilde{G_R}$, we iterate over each scene $i$ in the database. 
    \item For every scene $i$, the corresponding latent $Z_i = \{Z_i^1, \ldots Z_i^{n_{F_i}}\}$ where $n_{F_i}$ is the number of furniture items in scene $i$.
    \item Since $Z_i$ was sampled using our approximate posterior $q_{\phi}(Z \mid G_i, T_i, n_{F_i})$ we solve \eqref{eq: optimal ordering} using the FAQ algorithm and the Graph prior $p_{\theta''}(Z \mid \tilde{G_R}, \pi, T_i, n_{F_i})$\footnote{Recall, given any ordering of the latent variables, $\pi$, the Graph Prior simulates and auto-regressive model based on $\pi$. See \eqref{eq. graph prior based on ordering}} to find the optimal ordering $\pi^*$. Here $\tilde{G_R}$ is the given empty room layout.
    \item We then evaluate the likelihood of $Z_i$ under our prior and optimal ordering $\pi^*$
    \item Finally we choose the top $3$ scenes which have the highest likelihood under the prior as the closest "match". In other words, these latent codes are very likely under the prior for room layout $\tilde{G_R}$ and thus would result in good designs when passed through our GNN decoder. 
    \end{itemize}
\end{enumerate}

In Figure \ref{fig: database retrieval} we showed multiple retrieval results for a library, here we give two more examples a bedroom and a living room. Notice that the retrieved scenes can have very different room shape compared to $\tilde{G}_R$ however the furniture arrangements can still look plausible in $\tilde{G}_R$.

\subsection{Qualitative comparison of our method with ATISS and baselines}
\begin{figure}[h]
	\centering
	\includegraphics[width=0.85\linewidth]{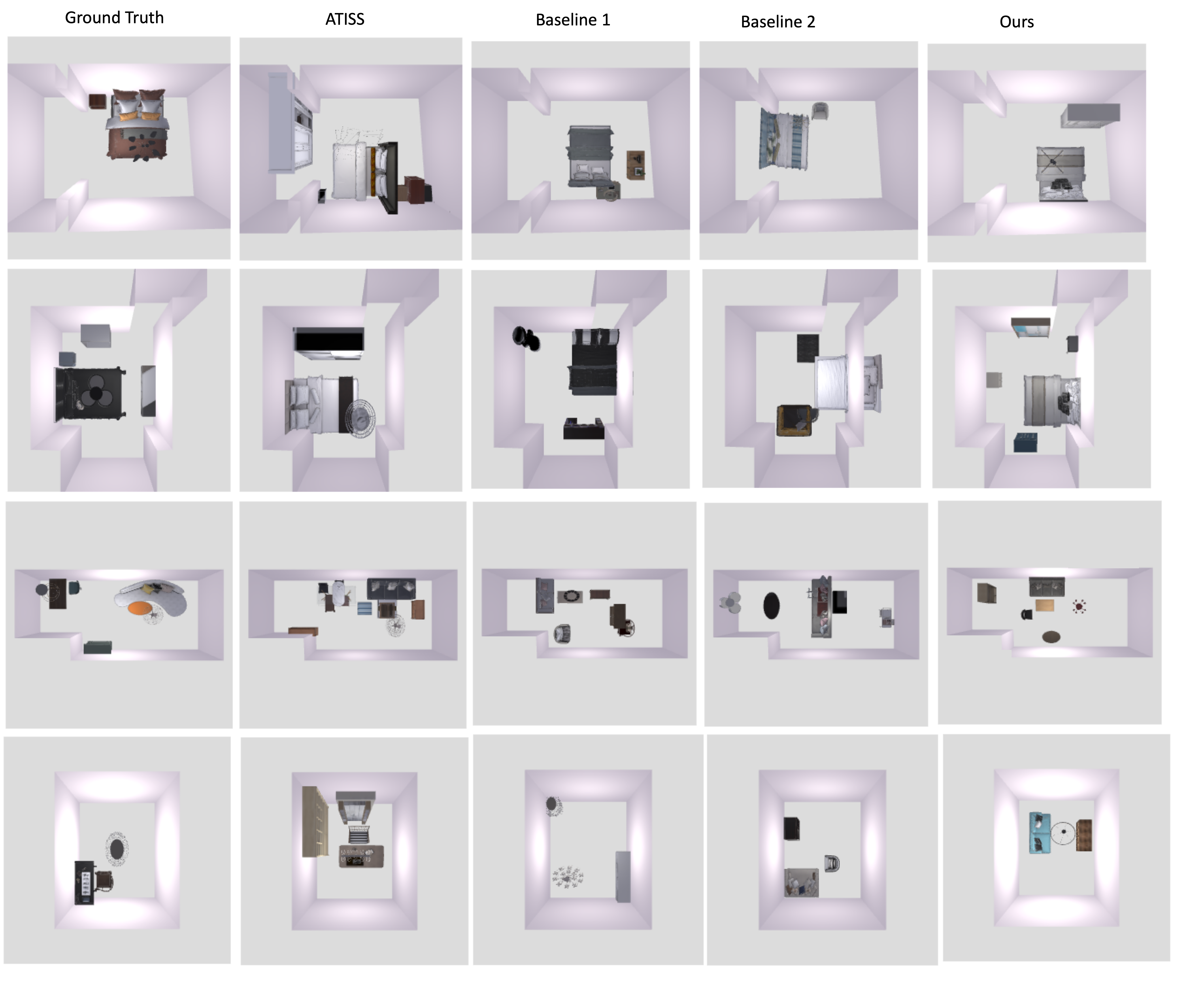}
	\caption{\textit{More results for Qualitative comparison of our method with ATISS and baselines. Row 1,2 are bedrooms. Row 3 is a living room and Row 4 is a library.}}	
	\label{fig: more qualitative comparison}
\end{figure}

In this subsection, we show more Examples for Figure \ref{fig: qualitative comparison} in Figure \ref{fig: more qualitative comparison}.

\subsection{More results for scene editing}
\begin{figure}[h]
	\centering
	\includegraphics[width=\linewidth]{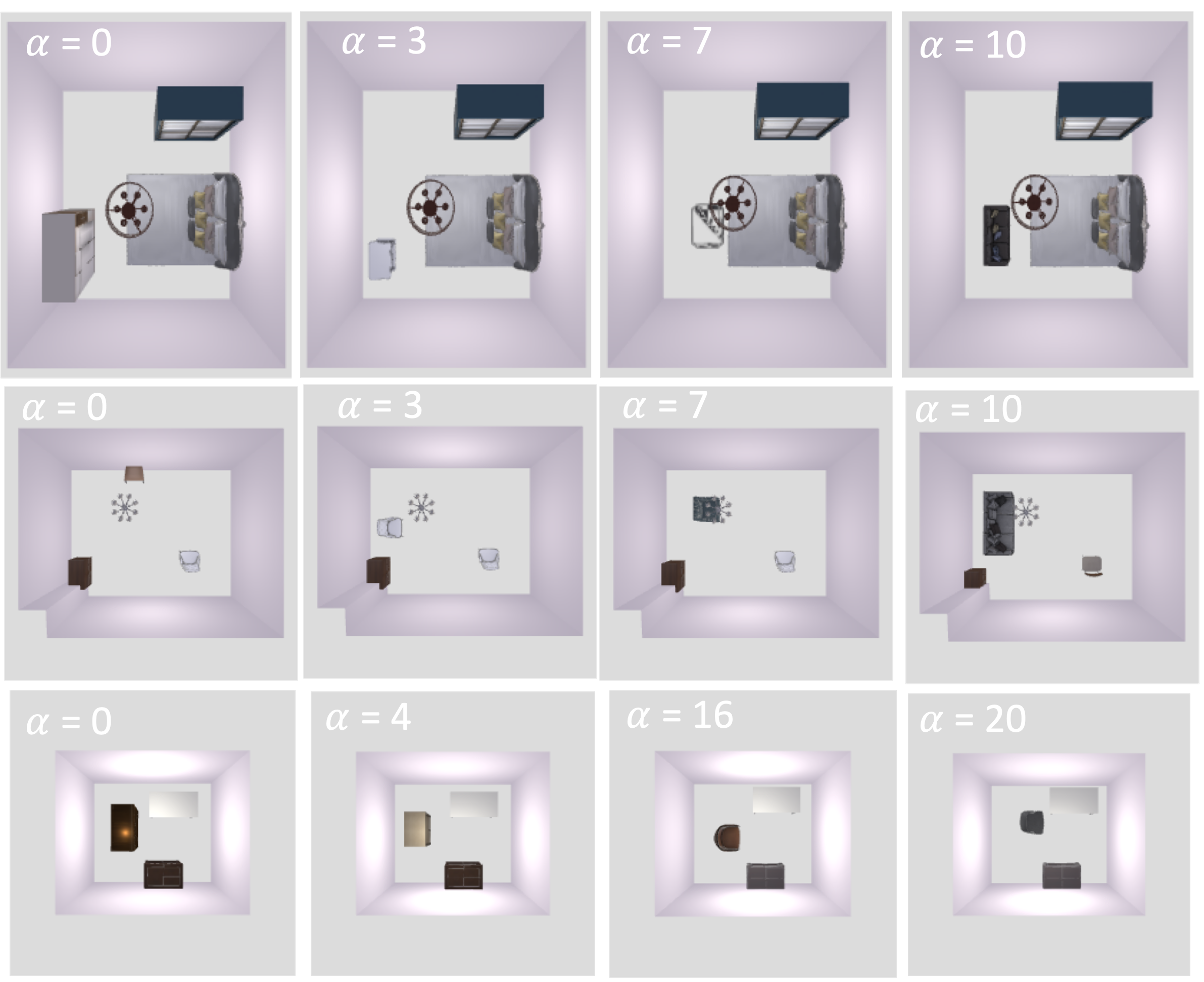}
   \caption{\textit{\textbf{Scene Editing.} Col 1 is a scene generated by our model. In row 1, we morph the bottom-left cabinet into a sofa by changing the $\alpha$ parameter as explained in text (\S\ref{sec. experiments}). In row 2 we morph the yellow chair in the top-center into a sofa. In row 3, we morph the cabinet (marked with a yellow spot) into a chair.}}
	\label{fig: more scene editing}
\end{figure}

In this subsection, we show more Examples for Figure \ref{Figure: Latent space traversals} in Figure \ref{fig: more scene editing}.

\newpage
\subsection{Analyzing matched furniture nodes for the trained autoregressive prior}
In this subsection we analyze the following question - After training, what is the category of the furniture's latent, taken from an encoded scene using the GNN encoder, that gets matched to the first $Z^1$ sampled using our autoregressive prior? 

To answer this we carry out the following steps,
\begin{itemize}
    \item Iterate over the scenes in the test set.
    \item Ever scene is a tuple consisting of the attributed scene graph, the room type and the number of furniture's in the room $(G, T, n_F)$. For each scene, 
    \begin{itemize}
        \item Pass it through the trained GNN encoder to obtain the parameters for $q_{\phi}(Z \mid G, T, n_F)$. Here $Z := \{Z^1, Z^2, \ldots Z^{n_F}\}$ is the set of latent variables corresponding to the $n_F$ furniture items to be placed in the room.
        \item Next solve \eqref{eq: optimal ordering} using the FAQ algorithm and the Graph prior $p_{\theta''}(Z \mid G_R, \pi, T, n_F)$\footnote{Recall, given any ordering of the latent variables, $\pi$, the Graph Prior simulates and auto-regressive model based on $\pi$. See \eqref{eq. graph prior based on ordering}} to find the optimal ordering $\pi^*$. Here $G_R$ refers to the room layout graph derived from the input scene graph. 
        \item This $\pi^*$ is the optimal assignment between $Z^i$'s obtained from the approximate posterior $q_{\phi}(Z \mid G, T, n_F)$ and the sequential latent nodes sampled using the auto-regressive prior (See \eqref{eq: autoregressive model}).
    \end{itemize}
    \item Since each $Z^i$ corresponds to a furniture item, we analyze the frequencies (over the test set) with which a certain category is mapped to the first latent sampled by the auto-regressive prior. This can give insights into what the model has learnt and whether the prior's modelling of furniture placement in indoor scenes correlates with how interior designers begin planning room layouts. In Table \ref{Table: matched furniture nodes} we present these results. We found that the first bedroom item matched by our model is a nightstand/bed/light with probability $0.91$ and the first living room item matched is a coffee-table/sofa/light with probability $0.76$. This is interesting since human designers often start planning with these items.
\end{itemize}

\begin{table}[h]
\caption{\textit{Frequency of categories mapped to the first latent sampled by the auto-regressive prior}}
    \centering
    \begin{tabular}{l|c|c}
        \toprule
        \textbf{Categories} & \textbf{Bedroom} & \textbf{Livingroom} \\
        \toprule
        Bed & 0.30 & 0.0 \\
        Light & 0.19 & 0.19 \\
        Night-stand & 0.43 & 0.0 \\
        Chair & 0.02 & 0.11 \\
        Sofa & 0.01 & 0.20 \\
        Table & 0.04 & 0.10 \\
        Pier & 0.01 & 0.01 \\
        Coffee-Table & 0.0 & 0.39 \\
        \bottomrule
    \end{tabular}
    \label{Table: matched furniture nodes}
\end{table}